\documentclass[runningheads]{llncs}

\usepackage{eccv}
\usepackage{eccvabbrv}
\usepackage{graphicx}
\usepackage{booktabs}
\usepackage[accsupp]{axessibility}
\usepackage{enumitem}
\usepackage{indentfirst}
\usepackage{makecell}
\usepackage{multirow}
\usepackage{pgf}
\usepackage{pifont}
\usepackage{tabularx}
\usepackage{tikz}
\usepackage{hyperref}
\usepackage{orcidlink}

\newcommand{\model}{$\rm R^2$-Tuning}
\newcommand{\modelbold}{$\mathbf{R^2}$\textbf{-Tuning}}
\newcommand{\block}{$\rm R^2$\:Block}

\newcommand{\paragraphbf}[1]{\paragraph{\rm\textbf{#1}}}

\newcommand{\cmark}{\ding{51}}
\newcommand{\xmark}{\ding{55}}
\newcommand{\gmark}{\color[HTML]{00B050}\ding{51}}
\newcommand{\omark}{\color[HTML]{F79646}\ding{51}}
\newcommand{\rmark}{\color[HTML]{FF0000}\ding{55}}

\definecolor{clipscolor}{rgb}{0.906,0.702,0.886}
\definecolor{clipbcolor}{rgb}{0.588,0.871,0.784}
\definecolor{sfcolor}{rgb}{0.694,0.886,0.988}
\definecolor{panncolor}{rgb}{0.961,0.871,0.702}

\newcommand{\clipsdot}{\raisebox{-0.65pt}{
\begin{tikzpicture}
\fill[clipscolor] (0,0) circle (.8ex);
\draw[black, line width=0.5pt] (0,0) circle (.8ex);
\end{tikzpicture}}}

\newcommand{\clipbdot}{\raisebox{-0.65pt}{
\begin{tikzpicture}
\fill[clipbcolor] (0,0) circle (.8ex);
\draw[black, line width=0.5pt] (0,0) circle (.8ex);
\end{tikzpicture}}}

\newcommand{\sfdot}{\raisebox{-0.65pt}{
\begin{tikzpicture}
\fill[sfcolor] (0,0) circle (.8ex);
\draw[black, line width=0.5pt] (0,0) circle (.8ex);
\end{tikzpicture}}}

\newcommand{\panndot}{\raisebox{-0.65pt}{
\begin{tikzpicture}
\fill[panncolor] (0,0) circle (.8ex);
\draw[black, line width=0.5pt] (0,0) circle (.8ex);
\end{tikzpicture}}}

\newcommand{\clipsfdot}{\raisebox{-0.65pt}{
\begin{tikzpicture}
\fill[clipscolor] (0,0) -- (270:.8ex) arc (270:90:.8ex) -- cycle;
\fill[sfcolor] (0,0) -- (90:.8ex) arc (90:-90:.8ex) -- cycle;
\draw[black, line width=0.5pt] (0,0) circle (.8ex);
\end{tikzpicture}}}

\newcommand{\vadot}{\raisebox{-0.65pt}{
\begin{tikzpicture}
\fill[sfcolor] (0,0) -- (270:.8ex) arc (270:90:.8ex) -- cycle;
\fill[panncolor] (0,0) -- (90:.8ex) arc (90:-90:.8ex) -- cycle;
\draw[black, line width=0.5pt] (0,0) circle (.8ex);
\end{tikzpicture}}}

\begin{document}

\title{\model: Efficient Image-to-Video Transfer Learning for Video Temporal Grounding}

\titlerunning{Reversed Recurrent Tuning}

\author{Ye~Liu\inst{1,3}$^*$\orcidlink{0000-0001-9597-0525} \and Jixuan~He\inst{2,3}\orcidlink{0009-0006-0543-4475} \and Wanhua~Li\inst{3}$^\dag$\orcidlink{0000-0002-2730-0543} \and Junsik~Kim\inst{3}\orcidlink{0000-0003-2555-5232} \and Donglai~Wei\inst{4}\orcidlink{0000-0002-2329-5484} \and Hanspeter~Pfister\inst{3}\orcidlink{0000-0002-3620-2582} \and Chang~Wen~Chen\inst{1}$^\dag$\orcidlink{0000-0002-6720-234X}}

\authorrunning{Y.~Liu et al.}

\institute{$^1$\,The Hong Kong Polytechnic University\; $^2$\,Tsinghua University \\
$^3$\,Harvard University\; $^4$\,Boston College \\
\email{coco.ye.liu@connect.polyu.hk}\; \email{hejixuan2000@gmail.com}\; \email{wanhua@seas.harvard.edu}\; \email{mibastro@gmail.com}\; \email{donglai.wei@bc.edu}\; \email{pfister@seas.harvard.edu}\; \email{changwen.chen@polyu.edu.hk}}

\maketitle

\let\thefootnote\relax\footnotetext{$^*$\,Work done at Harvard University. $^\dag$\,Corresponding authors.}

\begin{abstract}
Video temporal grounding (VTG) is a fine-grained video understanding problem that aims to ground relevant clips in untrimmed videos given natural language queries. Most existing VTG models are built upon frame-wise final-layer CLIP features, aided by additional temporal backbones (\eg, SlowFast) with sophisticated temporal reasoning mechanisms. In this work, we claim that CLIP itself already shows great potential for fine-grained spatial-temporal modeling, as each layer offers distinct yet useful information under different granularity levels. Motivated by this, we propose \underline{R}eversed \underline{R}ecurrent \underline{Tuning} (\model), a parameter- and memory-efficient transfer learning framework for video temporal grounding. Our method learns a lightweight \block{} containing only $1.5\%$ of the total parameters to perform progressive spatial-temporal modeling. Starting from the last layer of CLIP, \block{} recurrently aggregates spatial features from earlier layers, then refines temporal correlation conditioning on the given query, resulting in a coarse-to-fine scheme. \model{} achieves state-of-the-art performance across three VTG tasks (\ie, moment retrieval, highlight detection, and video summarization) on six public benchmarks (\ie, QVHighlights, Charades-STA, Ego4D-NLQ, TACoS, YouTube Highlights, and TVSum) even without the additional backbone, demonstrating the significance and effectiveness of the proposed scheme. Our code is available at \url{https://github.com/yeliudev/R2-Tuning}.
\end{abstract}

\section{Introduction}\label{sec:intro}

\begin{figure}[t]
\begin{minipage}{0.545\textwidth}
\centering
\includegraphics[height=2.7cm]{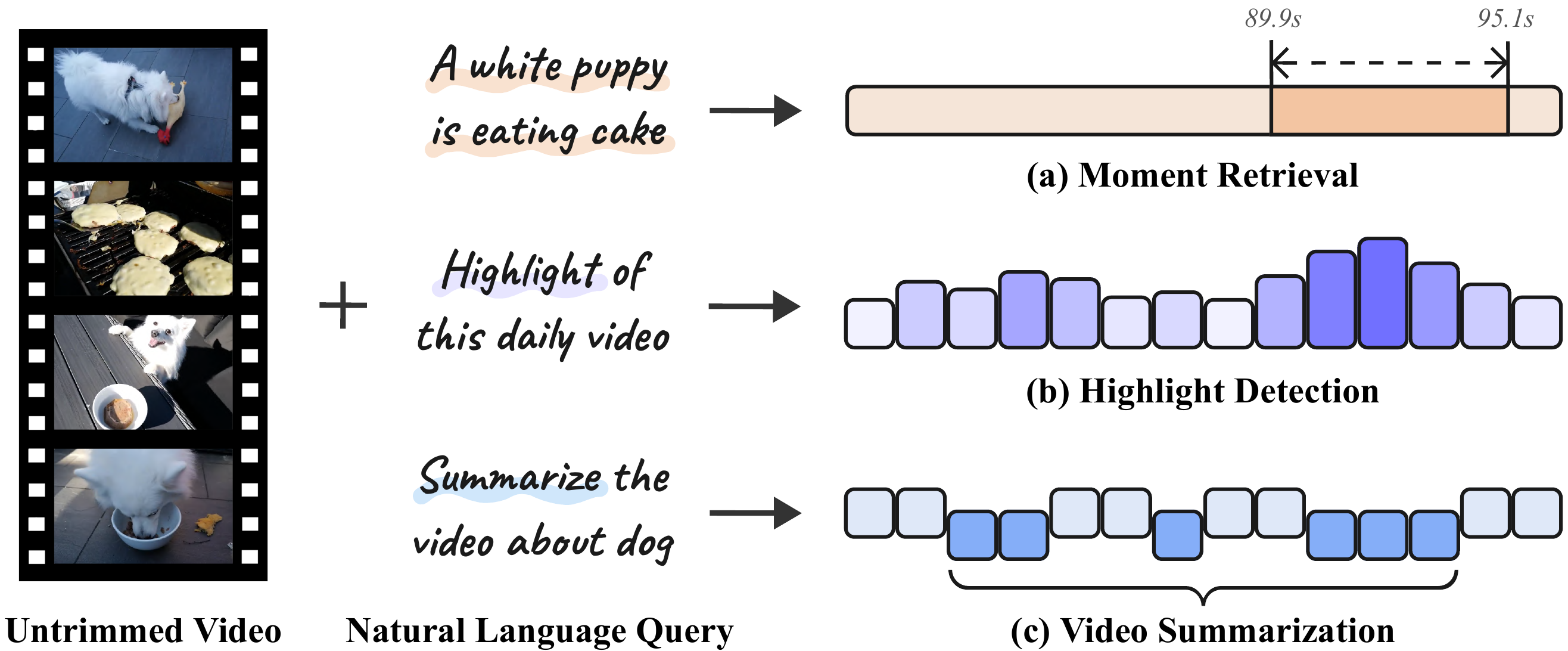}
\caption{Video temporal grounding (VTG) contains three video-language understanding problems, \ie, moment retrieval (MR), highlight detection (HD), and video summarization (VS).}
\label{fig:task}
\end{minipage}
\hfill
\begin{minipage}{0.425\textwidth}
\centering
\resizebox{!}{2.7cm}{\input{figures/cliponly.pgf}}
\caption{Moment retrieval mAP with different backbones on QVHighlights \texttt{val} split. CLIP's potential of temporal modeling was not fully exploited.}
\label{fig:cliponly}
\end{minipage}
\end{figure}

Video is becoming the major content media in our daily lives. The variety of video domains has extended beyond daily activities \cite{carreira2017quo,lei2021qvhighlights} but broader scenes such as egocentric \cite{grauman2022ego4d} and surveillance \cite{yuan2023ucf}. They maintain large information capacities within multi-granularities, and thus can convey both high-level context and low-level details effectively.

In the content production and consumption industry, such a flourishing is accompanied by the increasing demand for browsing untrimmed videos efficiently with different user interests. This derives the topic of video temporal grounding (VTG) \cite{lin2023univtg}, which is a fine-grained video-language understanding problem aiming to ground video clips conditioning on natural language queries. As shown in Figure~\ref{fig:task}, VTG can be disentangled into three tasks based on different output formats, \ie, moment retrieval (MR) \cite{gao2017tall,lei2021qvhighlights,liu2022umt,grauman2022ego4d} for regressing start-end timestamps, highlight detection (HD) \cite{sun2014ranking,lei2021qvhighlights} for predicting frame-level saliency curves, and video summarization (VS) \cite{gygli2014creating,song2015tvsum,apostolidis2021video} for classifying disjoint shots.

Encouraged by the recent success in adopting vision-language models (VLMs) for video understanding \cite{lin2022frozen,ni2022expanding,ju2022prompting,luo2022clip4clip,bain2022clip,huang2023vop}, most existing VTG methods \cite{lei2021qvhighlights,liu2022umt,xu2023mh,moon2023query,jang2023knowing,moon2023correlation,sun2024tr} are built upon frame-wise final-layer features from CLIP \cite{radford2021learning}. However, due to CLIP's misaligned pre-training objective (image-text contrast), these methods fail to capture temporal correlations well. As a feasible compromise, an additional backbone (\eg, SlowFast \cite{feichtenhofer2019slowfast}) is incorporated to complement the temporal information, followed by carefully designed modules such as text-guided queries \cite{liu2022umt}, dynamic anchors \cite{moon2023query}, and event reasoning \cite{jang2023knowing}.

\begin{figure}[t]
\centering
\includegraphics[width=\linewidth]{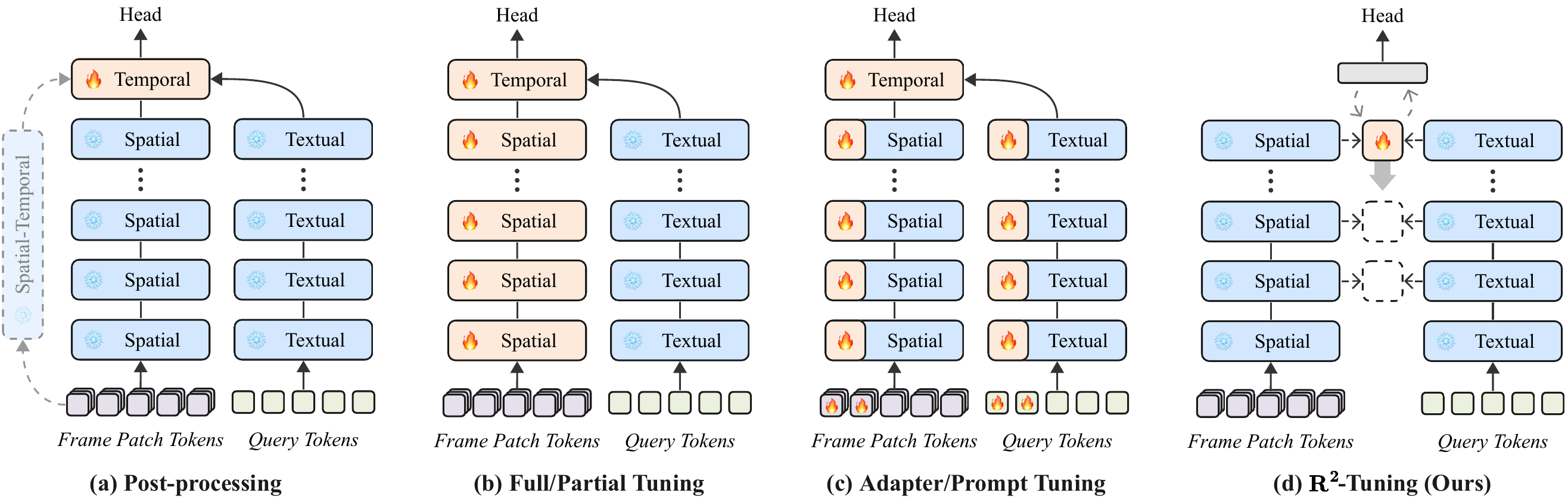}
\caption{Different architectural designs for CLIP-based image-to-video transfer learning. The gray rectangle in (d) denotes the progressively refined spatial-temporal features.}
\label{fig:design}
\end{figure}

We refer to the paradigm above as a \textit{post-processing} scheme (shown in Figure.~\ref{fig:design}~(a)), whereas two natural drawbacks exist due to the sub-optimal design. First, leveraging two backbones with similar capabilities is unintuitive and inefficient during inference. A single model with both vision-text alignment and spatial-temporal modeling abilities is more preferred. Second, queries for VTG could be of different granularities from coarse (\eg, \textit{the family is traveling}) to fine (\eg, \textit{when did I take the golf club from the man with white hair}). Leveraging only frame-wise \& final-layer features is not granularity flexible, as it would focus more on high-level scene transitions while overlooking low-level details. Preliminary experiments in Figure~\ref{fig:cliponly} also demonstrate that the potential of spatial-temporal modeling for CLIP is not fully exploited by existing methods. Some recent works \cite{yan2023unloc,ju2022prompting,huang2023vop} tried to tackle the first problem by fine-tuning (part of) the CLIP encoders (Figure~\ref{fig:design}~(b)~and~(c)), but they are all inefficient in terms of data, learnable parameters, or memory.

This paper intends to answer the research question: \textit{how to efficiently transfer an image-language foundation model to video temporal grounding?} We consider the above question by exploring two aspects: \textit{efficiency} and \textit{granularity flexibility}. To address these issues, a novel image-to-video transfer learning framework called \underline{R}eversed \underline{R}ecurrent \underline{Tuning} (\model) is proposed for fine-grained understanding on untrimmed videos. Our insight is that multi-layer CLIP features offer distinct yet useful information, while their integration should be tailored to the downstream task. As illustrated in Figure~\ref{fig:design}~(d), based on a frozen CLIP \cite{radford2021learning}, our method learns a side-block (\block) containing only $1.5\%$ of the total parameters to perform spatial-temporal modeling. \block{} is recurrently attached to the last few layers of CLIP encoders from back to front, performing \textit{query-modulated spatial pooling} and \textit{recurrent temporal refinement} from coarse to fine. Aside from the new architecture, we also introduce video-level and layer-wise constraints to calibrate the granularities of visual and text encoders. During training, gradient flows do not pass through the CLIP encoders, thus our scheme is both parameter- and memory-efficient. It is also granularity-flexible as \block{} can adaptively control the spatial pooling strategy conditioning on queries.

We conduct extensive experiments across three VTG tasks on six public benchmarks, including QVHighlights \cite{lei2021qvhighlights}, Ego4D-NLQ \cite{grauman2022ego4d}, Charades-STA \cite{gao2017tall}, TACoS \cite{gao2017tall}, YouTube Highlights \cite{sun2014ranking}, and TVSum \cite{song2015tvsum}. Without bells and whistles, \model{} achieves more than $3$ MR mAP gain on QVHighlights \texttt{test} split compared with $4\times$ heavier counterparts \cite{moon2023correlation} with additional temporal backbones and carefully designed post-processing modules. Further analysis also shows that our method can better handle multi-granularity information. Overall, our contributions are summarized as: (1) We introduced \model, a novel image-to-video transfer learning framework tailored for video temporal grounding. (2) We designed two effective strategies, \ie, \textit{query-modulated spatial pooling} and \textit{recurrent temporal refinement}, to model spatial-temporal information from coarse to fine. (3) To calibrate the granularities of CLIP visual and text encoders, we further introduce video-level and layer-wise contrastive constraints to distill distinct information from each layer. (4) Extensive experiments across three tasks on six benchmarks demonstrate the significance and effectiveness of \model.

\section{Related Work}

\paragraphbf{CLIP for Video Understanding}

With powerful transfer learning abilities, CLIP \cite{radford2021learning} has been widely used for many image-language tasks \cite{liu2022prompt,jiang2023clip}, and how to extend them to video understanding is an emerging and crucial research topic. Early attempts have been made to transfer CLIP to trimmed videos for action recognition \cite{lin2022frozen,ni2022expanding,rasheed2023fine} and video-text retrieval \cite{ju2022prompting,luo2022clip4clip,bain2022clip,huang2023vop}. Specifically, Efficient-Prompt \cite{ju2022prompting} introduces learnable prompt tokens to the text encoder, and ViFi-CLIP \cite{rasheed2023fine} further adopts learnable prompts at the visual encoder. Other works also tried to learn external modules designed for temporal modeling. EVL \cite{lin2022frozen} proposes a temporal decoder in parallel with the main network. CLIP4Clip \cite{luo2022clip4clip} investigated multiple temporal fusion strategies for video frames. Nevertheless, how to effectively transfer CLIP to untrimmed videos remains unexplored. To our best knowledge, \model{} is the first solution for fine-grained video temporal grounding via memory-efficient transfer learning.

\paragraphbf{Video Temporal Grounding}

Video temporal grounding (VTG) \cite{lin2023univtg} is a bundle of video understanding problems including moment retrieval (MR) \cite{gao2017tall,lei2021qvhighlights,liu2022umt,xu2023mh,moon2023query,jang2023knowing,yan2023unloc,yuan2019semantic,nan2021interventional,wang2022negative}, highlight detection (HD) \cite{sun2014ranking,hong2020mini,xu2021cross,badamdorj2021joint}, and video summarization (VS) \cite{gygli2014creating,song2015tvsum}. Although these tasks are closely related, they have not been jointly studied until recently. QVHighlights \cite{lei2021qvhighlights} is the first dataset that supports both MR and HD, breaking the moment distribution bias in previous MR datasets \cite{regneri2013grounding,gao2017tall}. The authors also proposed Moment-DETR, a strong baseline method for MR/HD. Most following works were built upon the Moment-DETR framework with specially designed modules. Notably, UMT \cite{liu2022umt} tackles flexible MR and HD through a unified multi-modal architecture. QD-DETR \cite{moon2023query} benefits from query-dependent video representations and dynamic anchors. EaTR \cite{jang2023knowing} incorporates moment reasoning to provide reliable referential search areas for moment queries. Nevertheless, their design space is still within pre-extracted CLIP \cite{radford2021learning} and SlowFast \cite{feichtenhofer2019slowfast} features. A recent work \cite{yan2023unloc} revealed the possibility of fine-tuning CLIP for MR, while their solution (post pre-training) is extremely resource-consuming. Compared with existing works, \model{} can effectively learn strong temporal modeling abilities without any pre-training on videos.

\paragraphbf{Parameter-Efficient Transfer Learning}

Foundation models \cite{radford2021learning,touvron2023llama} have achieved remarkable successes in vision and language understanding. Efforts have been made to efficiently transfer knowledge from these models to new scenarios. One line of research is to conduct context-based tuning (\eg, prompt-tuning \cite{zhou2022learning,zhou2022conditional,jia2022visual}), which does not modify the model architectures but incorporates learnable embeddings in the inputs. Another paradigm is introducing an extra lightweight adapter to the original model, and keeping the rest of the parameters frozen \cite{pan2022st}. Recent works also consider memory-efficient transfer learning (METL) that introduces parameters whose gradient flows do not go through the main model \cite{sung2022lst,lin2022frozen,qing2023disentangling}. \model{} is also a METL framework, where we adopt a novel recurrent tuning strategy to ensure parameter- and memory-efficient.

\begin{figure}[t]
\centering
\includegraphics[width=\linewidth]{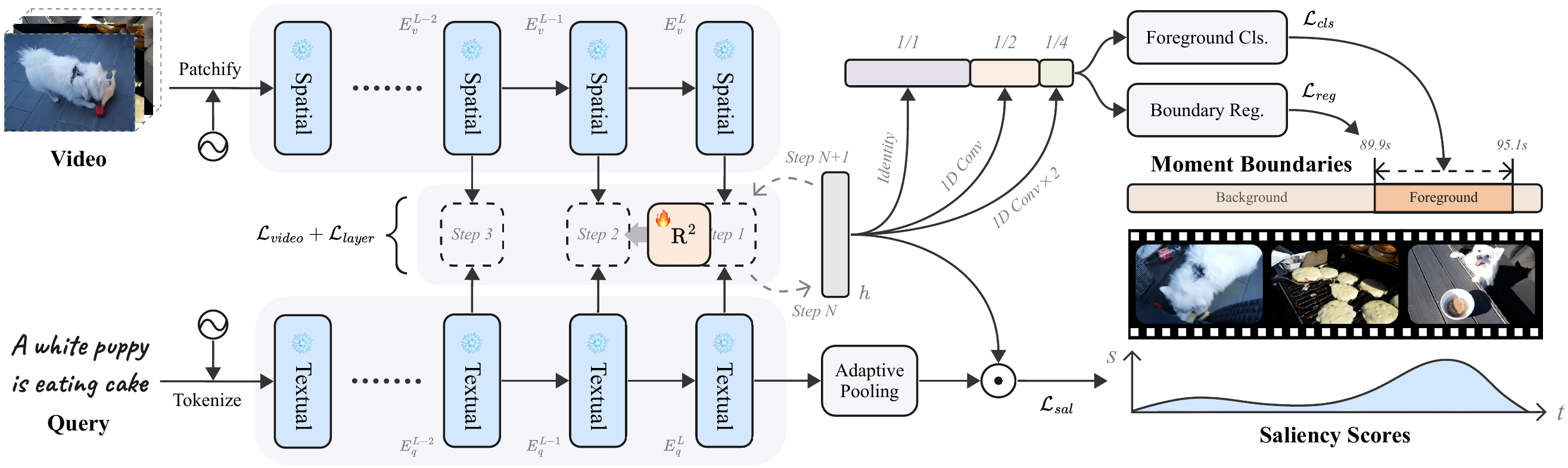}
\caption{Overall architecture of our framework. The input video and query are first encoded by frozen CLIP \cite{radford2021learning} encoders. Their multi-layer outputs are then recurrently fused and refined by a learnable \block{} to construct spatial-temporal representations $h$, which would be scaled up/down to construct a temporal feature pyramid, followed by three heads for MR, HD, and VS, respectively.}
\label{fig:model}
\end{figure}

\section{Methodology}

\subsection{Problem Formulation}

Given a video $V = \{v_i\}_{i=1}^{T}$ and a natural language query $Q = \{q_i\}_{i=1}^{L}$, where $T$ and $L$ are the numbers of video frames and text tokens, VTG aims to densely predict a set of labels $(b_i, s_i, f_i)$ for each frame, defined as follows:

\begin{itemize}
\item \textbf{Moment Retrieval} is to find the most relevant moments (\ie sets of consecutive frames) in $V$ according to $Q$, so that $b_i = [b_i^s, b_i^e] \in \mathbb{R}^2$ represents the temporal displacements from frame $v_i$ to the start and end timestamps of the nearest target moment.
\item \textbf{Highlight Detection} requires estimating frame-level relevancies between $V$ and $Q$, thus $s_i \in [0, 1]$ is a continuous saliency score denoting to what extent frame $v_i$ is semantically aligned with $Q$.
\item \textbf{Video Summarization} aims to select a subset of video frames according to $Q$ to form a concise summary, therefore, $f_i \in \{0, 1\}$ is a binary score indicating whether frame $v_i$ belongs to the summary.
\end{itemize}

\subsection{Overview}

Figure~\ref{fig:model} shows an overview of the proposed framework. Our model derives from a pre-trained and frozen CLIP \cite{radford2021learning} with ViT \cite{dosovitskiy2020image} backbone, which has a two-stream architecture for spatial and textual encoding, respectively. A learnable $\rm R^2$ block is iteratively attached to the last $K$ encoder layers to refine the temporal correlation. The resulting features are then scaled up/down to build a feature pyramid, followed by three heads to decode the task-specific predictions.

Specifically, the input video $V$ and query $Q$ are first tokenized into frame patches and word tokens, then sent into the visual encoder $E_v$ and query encoder $E_q$. The encoded visual and textual features can be denoted as $e_v \in \mathbb{R}^{B \times N \times T \times (P + 1) \times D_v}$ and $e_q \in \mathbb{R}^{B \times N \times L \times D_q}$, where $B$, $N$, $T$, $P$, $L$, $D_v$ and $D_q$ indicate batch size, number of encoder layers, number of video frames, number of patches per frame, number of query tokens, and the dimensions of visual/query features, respectively. These features are recurrently fused and refined by $\rm R^2$ block to construct spatial-temporal features $h \in \mathbb{R}^{B \times T \times C}$, in which each token preserves the $C$-dimensional features for a frame. This process will be introduced in detail in Section.~\ref{sec:r2t}.

\begin{figure}[t]
\centering
\includegraphics[width=0.55\linewidth]{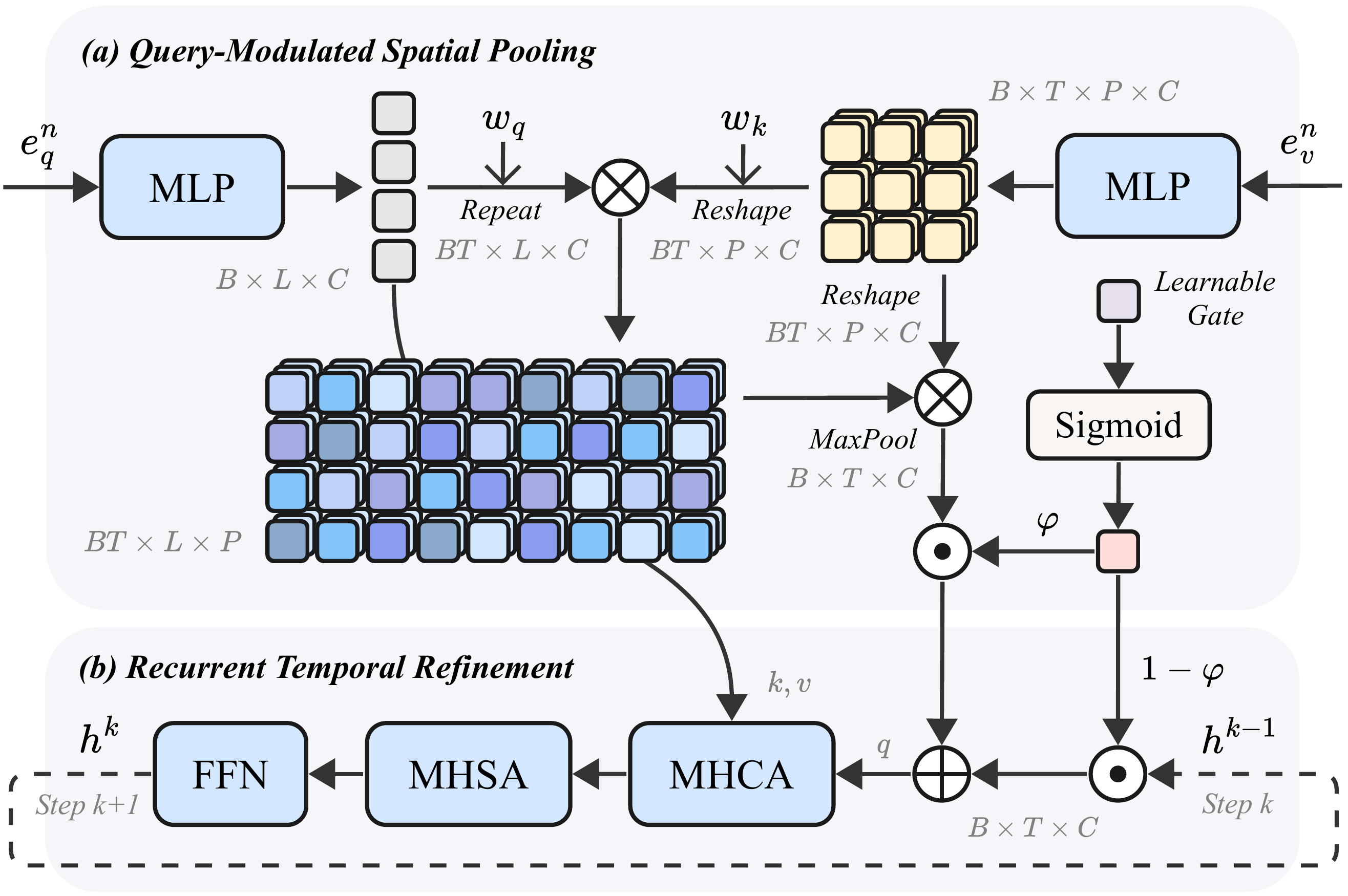}
\caption{Detailed architecture of the \block. It can be split into two parts: a) \textit{query-modulated spatial pooling}, and b) \textit{recurrent temporal refinement}. Note that the \texttt{[CLS]} tokens of visual features are omitted for clarity.}
\label{fig:block}
\end{figure}

\subsection{Reversed Recurrent Tuning}\label{sec:r2t}

Given pre-trained visual and query encoders, existing works \cite{lei2021qvhighlights,liu2022umt,lin2023univtg,moon2023query,jang2023knowing} merely take features from the last hidden layer, which is sub-optimal due to the limited information. In this work, we propose \model{} to exploit the potential of spatial-temporal modeling based on multi-layer CLIP features. Our scheme learns a lightweight \block{} which maintains a zero-initialized hidden state $h \in \mathbb{R}^{B \times T \times C}$ as frame-level spatial-temporal features. The \block{} is recurrently attached to the last $K$ layers of CLIP encoders from back to front to refine the hidden state $h$ for $K$ steps, with each step defined as follows:
\begin{gather}
h^k = \mathcal{F}_\theta(e_v^n, e_q^n, h^{k-1})
\end{gather}
Here, $\mathcal{F}_\theta$ is the refinement operation parameterized by $\theta$. $k \in [1, K]$ is the index of current step. $n = N - k + 1$ denotes the CLIP layer index for step $k$, thus $e_v^n$ and $e_q^n$ are visual and query features from the $n$-th CLIP encoder layer. The refinement operation $\mathcal{F}_\theta$ can be decomposed into two parts: 1) \textit{query-modulated spatial pooling}, and 2) \textit{recurrent temporal refinement}. Details are discussed as follows.

\paragraphbf{Query-Modulated Spatial Pooling}

Figure~\ref{fig:block}~(a) presents the data flow of query-modulated spatial pooling. The goal of this process is to adaptively pool spatial features from patch-level representations $e_v^n$ to a single token $e_{pool}^n$ conditioning on the query $e_q^n$. We first adopt two separate MLPs to map $e_v^n$ and $e_q^n$ into the same space:
\begin{align}
\hat{e}_v^n &= {\rm MLP}(e_v^n) \in \mathbb{R}^{B \times T \times (P + 1) \times C} \\
\hat{e}_q^n &= {\rm MLP}(e_q^n) \in \mathbb{R}^{B \times L \times C}
\end{align}
Here, $C$ is the reduced hidden size. We further align the shape of $\hat{e}_v^n$ and $\hat{e}_q^n$ by merging dimension $T$ into $B$ and repeat $\hat{e}_q^n$ for $T$ times in an interleaved style, resulting $\hat{e}_v^n \in \mathbb{R}^{(B \times T) \times (P + 1) \times C}$ and $\hat{e}_q^n \in \mathbb{R}^{(B \times T) \times L \times C}$. Then, we compute the similarities for patch-token pairs using normalized Embedded Gaussian \cite{vaswani2017attention}:
\begin{gather}\label{eq:attention}
a = {\rm softmax}(\frac{(w_q\hat{e}_q^n)^\top w_v\hat{e}_v^n}{\sqrt{C}}) \in \mathbb{R}^{(B \times T) \times L \times P}
\end{gather}
Here, $w_v$ and $w_q$ are learnable matrices for projecting features. This operation derives from cross-attention but discards the linear projection on \textit{value}. We then multiply the similarities $a$ with visual features $\hat{e}_v^n$ to pool them into each token, such that tokens can interact with patches independently. After that, a max pooling along the $L$ dimension is applied:
\begin{gather}
e_{token}^n = {\rm MaxPooling}(a\hat{e}_v^n + \hat{e}_q^n) \in \mathbb{R}^{B \times T \times C}
\end{gather}
Here, $\hat{e}_q^n$ serves as residuals \cite{he2016deep} to stabilize training. Our visualizations in Figure~\ref{fig:attention} show that this strategy can guide the model to focus more on query-related regions. Finally, we combine $e_{token}^n$ with the \texttt{[CLS]} token to generate query-modulated spatial features:
\begin{gather}\label{eq:fuse}
e_{pool}^n = e_v^{n,0} + g^k \cdot e_{token}^n \in \mathbb{R}^{B \times T \times C}
\end{gather}
Here, $g^k \in (-1,1)$ is a zero-initialized learnable gate for step $k$ constrained by $\rm Tanh$. We allow negative values here to remove useless information from \texttt{[CLS]} tokens. $e_{pool}^n$ is then used to model temporal correlations and are refined in a recurrent manner.

\paragraphbf{Recurrent Temporal Refinement}

Figure~\ref{fig:block}~(b) displays the flow of recurrent temporal refinement. Specifically, the pooled visual features $e_{pool}^n$ is first fused with the hidden state $h^{k-1}$ from the previous step:
\begin{gather}
\hat{h}^{k-1} = \varphi^k \cdot e_{pool}^n + (1 - \varphi^k) \cdot h^{k-1}
\end{gather}
where $\varphi^k \in (0, 1)$ is the learnable gate for step $k$. Then, we adopt a sequence of standard multi-head cross-attention ($\hat{e}_q^n$ as $k$, $v$), multi-head self-attention, followed by a feed-forward network \cite{vaswani2017attention} to update the hidden state:
\begin{gather}
h^k = {\rm FFN}({\rm MHSA}({\rm MHCA}(\hat{h}^{k-1}, \hat{e}_q^n)))
\end{gather}
For each block, we adopt DropPath \cite{larsson2016fractalnet} with $p = 0.1$ to prevent overfitting. The query features $\hat{e}_q^n$ are utilized both in spatial and temporal modeling as guidance.

\paragraphbf{Granularity Calibration}

The refinement process above is based on the assumption that visual and query features from the same layer of CLIP \cite{radford2021learning} are well aligned at the same granularity level. However, this cannot be guaranteed as the two encoders are learned in isolation during pre-training. Therefore, we need to add manual constraints to calibrate their granularities.

We apply a \textit{video-level constraint} and a \textit{layer-wise constraint} by designing two contrastive losses. We first denote $e_v^0 \in \mathbb{R}^{B \times K \times T \times C}$ as the features of $K$ \texttt{[CLS]} tokens from Eq.~\ref{eq:fuse}. Then, we select all positive frames (\ie, frames within $[b_i^s, b_i^e]$ or with $s_i$ or $f_i$ higher than a threshold) from $e_v^0$ and perform average pooling to obtain video-level representations $\widetilde{e}_v \in \mathbb{R}^{B \times K \times C}$.
\begin{gather}
\widetilde{e}_v = {\rm AvgPool}(\{e_v^{0,i}\}_{i \in \Omega})
\end{gather}
Here, $\Omega$ is the set of positive frame indices. To obtain query-level representations $\widetilde{e}_q \in \mathbb{R}^{B \times K \times C}$, a token-wise adaptive pooling is applied following previous work \cite{lin2023univtg}. We adopt InfoNCE loss \cite{oord2018representation} along two dimensions to calculate the video-level and layer-wise constraints:
\begin{align}
\mathcal{L}_{video} &= \lambda_{video}\frac{1}{B}\sum_{b \in B} {\rm InfoNCE}(\widetilde{e}_v^b, \widetilde{e}_q^b) \\
\mathcal{L}_{layer} &= \lambda_{layer}\frac{1}{K}\sum_{k \in K} {\rm InfoNCE}(\widetilde{e}_v^k, \widetilde{e}_q^k)
\end{align}
Here, $\mathcal{L}_{video}$ performs contrast among samples in the same batch (diversifying features are diverse among moment-query pairs), and averages the loss across $K$ layers. It also implicitly models temporal correlations between videos and queries. $\mathcal{L}_{layer}$ performs contrast among layers and averages across the batch. $\lambda_{video}$ and $\lambda_{layer}$ serve as re-weighting terms for the losses. A fixed temperature parameter of $0.07$ is used.

\subsection{Prediction Heads}

After refining spatial-temporal features $h$, a temporal feature pyramid is constructed by applying 1D convolutions with $stride = 2$ on $h$. The number of convolutions used for each level is subjected to $l - 1$, where $l$ is the level index starting from $1$. We concatenate features from all levels by the temporal dimension to form $\hat{h}$ and predict once in all heads. Following a similar but simplified design from \cite{lin2023univtg}, we adopt three heads for video temporal grounding, described in detail as follows.

\begin{table}[t]
\setlength{\tabcolsep}{1.05pt}
\fontsize{6.5pt}{7.5pt}\selectfont
\caption{Video moment retrieval (MR) and highlight detection (HD) results on QVHighlights \texttt{test} split. Note that \cite{jang2023knowing} and \cite{yan2023unloc} only reported their results on \texttt{val} split. \hspace{-0.5mm}\protect\clipsdot\hspace{0.5mm} and \hspace{-0.5mm}\protect\clipbdot\hspace{0.5mm} are CLIP-B/32 and B/16 \cite{radford2021learning}, \hspace{-0.5mm}\protect\sfdot\hspace{0.5mm} denotes SlowFast R-50 \cite{feichtenhofer2019slowfast}, and \hspace{-0.5mm}\protect\panndot\hspace{0.5mm} is PANN \cite{kong2020panns} for audio features extraction. $^*$ means estimated parameters. The best and second-best metrics are marked with \textbf{bold} and \underline{underline}, respectively.}
\label{tab:qvhighlights}
\centering
\begin{tabularx}{\linewidth}{lcccccccccccc}
\toprule
\multirow{4.15}{*}{\hspace{0.5cm}\textbf{Method}} & \multirowcell{4.15}{\textbf{Backbone}} & \multirowcell{4.1}{\textbf{Post} \\ \textbf{Pre-train}} & \multicolumn{6}{c}{\textbf{MR}} && \multicolumn{2}{c}{\textbf{HD}} & \multirowcell{4.15}{\textbf{\#Params}} \\
\cmidrule{4-9} \cmidrule{11-12}
&&& \multicolumn{2}{c}{R1} && \multicolumn{3}{c}{mAP} && \multicolumn{2}{c}{$\geqslant$ Very Good} \\
\cmidrule{4-5} \cmidrule{7-9} \cmidrule{11-12}
&&& @0.5 & @0.7 && @0.5 & @0.75 & Avg. && mAP & HIT@1 \\
\midrule
BeautyThumb \cite{song2016click} & \clipsdot \sfdot & \xmark & -- & -- && -- & -- & -- && 14.36 & 20.88 & -- \\
DVSE \cite{liu2015multi} & \clipsdot \sfdot & \xmark & -- & -- && -- & -- & -- && 18.75 & 21.79 & -- \\
MCN \cite{anne2017localizing} & \clipsdot \sfdot & \xmark & 11.41 & 2.72 && 24.94 & 8.22 & 10.67 && -- & -- & -- \\
CAL \cite{escorcia2019temporal} & \clipsdot \sfdot & \xmark & 25.49 & 11.54 && 23.40 & 7.65 & 9.89 && -- & -- & -- \\
XML \cite{lei2020tvr} & \clipsdot \sfdot & \xmark & 41.83 & 30.35 && 44.63 & 31.73 & 32.14 && 34.49 & 55.25 & -- \\
XML+ \cite{lei2021qvhighlights} & \clipsdot \sfdot & \xmark & 46.69 & 33.46 && 47.89 & 34.67 & 34.90 && 35.38 & 55.06 & -- \\
Moment-DETR \cite{lei2021qvhighlights} & \clipsdot \sfdot & \xmark & 52.89 & 33.02 && 54.82 & 29.40 & 30.73 && 35.69 & 55.60 & 4.8M \\
UMT \cite{liu2022umt} & \clipsdot \sfdot \panndot & \xmark & 56.23 & 41.18 && 53.83 & 37.01 & 36.12 && 38.18 & 59.99 & 14.9M \\
MomentDiff \cite{li2023momentdiff} & \clipsdot \sfdot \panndot & \xmark & 58.21 & 41.48 && 54.57 & 37.21 & 36.84 && -- & -- & -- \\
QD-DETR \cite{moon2023query} & \clipsdot \sfdot & \xmark & 62.40 & 44.98 && 62.52 & 39.88 & 39.86 && 38.94 & 62.40& 7.6M \\
MH-DETR \cite{xu2023mh} & \clipsdot \sfdot & \xmark & 60.05 & 42.48 && 60.75 & 38.13 & 38.38 && 38.22 & 60.51 & 8.2M \\
UniVTG \cite{lin2023univtg} & \clipsdot \sfdot & \xmark & 58.86 & 40.86 && 57.60 & 35.59 & 35.47 && 38.20 & 60.96 & 41.3M \\
TR-DETR \cite{sun2024tr} & \clipsdot \sfdot & \xmark & 64.66 & 48.96 && 63.98 & 43.73 & 42.62 && 39.91 & 63.42 & 7.9M \\
CG-DETR \cite{moon2023correlation} & \clipsdot \sfdot & \xmark & \underline{65.43} & 48.38 && \underline{64.51} & 42.77 & 42.86 && 40.33 & \underline{66.21} & 12.0M \\
\color{Gray} EaTR \cite{jang2023knowing} & \color{Gray} \clipsdot \sfdot & \color{Gray} \xmark & \color{Gray} 61.36 & \color{Gray} 45.79 && \color{Gray} 61.86 & \color{Gray} 41.91 & \color{Gray} 41.74 && \color{Gray} 37.15 & \color{Gray} 58.65 & \color{Gray} 9.0M \\
\midrule
Moment-DETR \cite{lei2021qvhighlights} & \clipsdot \sfdot & 236K ASR Cap. & 59.78 & 40.33 && 60.51 & 35.36 & 36.14 && 37.43 & 60.17 & 4.8M \\
UMT \cite{liu2022umt} & \clipsdot \sfdot \panndot & 236K ASR Cap. & 60.83 & 43.26 && 57.33 & 39.12 & 38.08 && 39.12 & 62.39 & 14.9M \\
UniVTG \cite{lin2023univtg} & \clipsdot \sfdot & 4.2M Corpus & \underline{65.43} & \textbf{50.06} && 64.06 & \underline{45.02} & \underline{43.63} && \underline{40.54} & \textbf{66.28} & 41.3M \\
\color{Gray} UnLoc \cite{yan2023unloc} & \color{Gray} \clipbdot & \color{Gray} 650K Videos & \color{Gray} 64.50 & \color{Gray} 48.80 && -- & -- & -- && -- & -- & \color{Gray} 87.9M$^*$ \\
\midrule
\modelbold{} (Ours) & \clipsdot & \xmark & \textbf{68.03} & \underline{49.35} && \textbf{69.04} & \textbf{47.56} & \textbf{46.17} && \textbf{40.75} & 64.20 & \textbf{2.7M} \\
\bottomrule
\end{tabularx}
\end{table}

\begin{table}[t]
\setlength{\tabcolsep}{1.25pt}
\fontsize{6.5pt}{7.5pt}\selectfont
\caption{Video moment retrieval results on Ego4D-NLQ, Charades-STA, and TACoS datasets. \hspace{-0.5mm}\protect\clipsfdot\hspace{0.5mm} means using both CLIP-B/32 and SlowFast R-50 as feature extractors. \hspace{-0.5mm}\protect\clipsdot\hspace{0.5mm} indicates using CLIP-B/32 only. The best and second-best metrics are marked with \textbf{bold} and \underline{underline}, respectively. Our method shows significant advantages in high-quality retrievals (R@0.7) even without a video backbone.}
\label{tab:moment}
\centering
\begin{tabularx}{\linewidth}{lcccccccccccccc}
\toprule
\multirow{2.65}{*}{\hspace{0.8cm}\textbf{Method}} & \multicolumn{4}{c}{\textbf{Ego4D-NLQ} \cite{grauman2022ego4d}} && \multicolumn{4}{c}{\textbf{Charades-STA} \cite{gao2017tall}} && \multicolumn{4}{c}{\textbf{TACoS} \cite{regneri2013grounding}} \\
\cmidrule{2-5} \cmidrule{7-10} \cmidrule{12-15}
& R@0.3 & R@0.5 & R@0.7 & mIoU && R@0.3 & R@0.5 & R@0.7 & mIoU && R@0.3 & R@0.5 & R@0.7 & mIoU \\
\midrule
\hspace{-1mm}\clipsfdot\hspace{0.1mm} 2D-TAN \cite{zhang2020learning} & 4.33 & 1.83 & 0.60 & 3.39 && 58.76 & 46.02 & 27.50 & 41.25 && 40.01 & 27.99 & 12.92 & 27.22 \\
\hspace{-1mm}\clipsfdot\hspace{0.1mm} VSLNet \cite{zhang2020span} & 4.54 & 2.40 & 1.01 & 3.54 && 60.30 & 42.69 & 24.14 & 41.58 && 35.54 & 23.54 & 13.15 & 24.99 \\
\hspace{-1mm}\clipsfdot\hspace{0.1mm} Moment-DETR \cite{lei2021qvhighlights} & 4.34 & 1.81 & 0.65 & 3.53 && 65.83 & 52.07 & 30.59 & 45.54 && 37.97 & 24.67 & 11.97 & 25.49 \\
\hspace{-1mm}\clipsfdot\hspace{0.1mm} UniVTG \cite{lin2023univtg} & \textbf{7.28} & \underline{3.95} & \underline{1.32} & \underline{4.91} && \underline{70.81} & \underline{58.01} & \underline{35.65} & \underline{50.10} && \textbf{51.44} & \underline{34.97} & \underline{17.35} & \underline{33.60} \\
\midrule
\hspace{-1mm}\clipsdot\hspace{0.1mm} \modelbold{} (Ours) & \underline{7.20} & \textbf{4.49} & \textbf{2.12} & \textbf{4.94} && \textbf{70.91} & \textbf{59.78} & \textbf{37.02} & \textbf{50.86} && \underline{49.71} & \textbf{38.72} & \textbf{25.12} & \textbf{35.92} \\
\bottomrule
\end{tabularx}
\end{table}

\paragraphbf{Foreground-Background Classification}

Given $\hat{h}$, a two-layer 1D convolution module with $(1 \times 3)$ kernels is adopted to predict $\hat{f}_i$ for each frame. It is optimized using a focal loss \cite{lin2017focal} with $\alpha = 0.9$ and $\gamma = 2.0$:
\begin{gather}
\mathcal{L}_{cls} = -\lambda_{cls}\alpha(1 - \hat{f}_i)^\gamma\log(\hat{f}_i)
\end{gather}
We find that focal loss perform better than binary cross-entropy when the numbers of foreground/background frames are severely imbalanced as noted in \cite{lin2023univtg}.

\paragraphbf{Boundary Regression}

Similar to foreground-background classification, boundary regression for moments is also realized by a two-layer 1D convolution module. The difference is that its output dimension is set to $2$ instead of $1$, representing the boundary displacements $[\hat{b}_i^s, \hat{b}_i^e]$ for start-end timestamps. This head is optimized by an L1 loss:
\begin{gather}
\mathcal{L}_{reg} = \lambda_{reg}(|b_i^s - \hat{b}_i^s| + |b_i^e - \hat{b}_i^e|)
\end{gather}
where $b_i^s$ and $b_i^e$ are the ground truths. This loss is only applied to frames inside the ground truth boundaries. We observe that using an L1 loss already works better than the combination of Smooth L1 Loss and GIoU Loss as in \cite{lin2023univtg}.

\paragraphbf{Saliency Prediction}

To obtain the saliency scores $\hat{s}_i$ for HD, we calculate the cosine similarities between the adaptive pooled query features at the last refinement step $\widetilde{e}_q^K$ and each token in the spatial-temporal features $h$:
\begin{gather}
\hat{s}_i = \frac{h_i^\top\widetilde{e}_q^K}{||h_i||_2||\widetilde{e}_q^K||_2}
\end{gather}
The training objective is applied through a contrastive loss between sampled positive frames (with index $p \in P$) and the adaptively pooled query $\widetilde{e}_q^K$:
\begin{gather}
\mathcal{L}_{sal} = -\lambda_{sal}\log\frac{\exp(\hat{s}_p/\tau)}{\exp(\hat{s}_p/\tau) + \sum_{i \in \Theta}\exp(\hat{s}_i/\tau)}
\end{gather}
Here, $\Theta$ is the set of frame indices where $s_i < s_p$, and $\tau$ is a fixed temperature parameter set to $0.07$.

\paragraphbf{Training \& Inference}

The whole model is jointly optimized using a sum of the five losses mentioned above. In practice, the loss weights are set as $\lambda_{video} = 0.1$, $\lambda_{layer} = 0.1$, $\lambda_{cls} = 1.0$, $\lambda_{reg} = 0.1$, and $\lambda_{sal} = 0.1$. During inference, we form an moment retrieval prediction by combining $f_i$ and $b_i$, \ie, calculating the start-end timestamps from frame index $i$ and boundary displacements $[b_i^s, b_i^e]$, while regarding $f_i$ as the confidence. NMS with IoU threshold $\theta_{IoU} = 0.7$ is applied to reduce duplicates. For highlight detection and video summarization, we directly use the frame-level output from saliency prediction $s_i$.

\section{Experiments}

\subsection{Datasets \& Evaluation Metrics}

We conduct experiments on six datasets covering various domains including daily vlogs \& news (QVHighlights \cite{lei2021qvhighlights}), in-door scenes (TACoS \cite{regneri2013grounding} and Charades-STA \cite{gao2017tall}), egocentric videos (Ego4D-NLQ \cite{grauman2022ego4d}), and sports (YouTube Highlights \cite{sun2014ranking} and TVSum \cite{song2015tvsum}). Details about the datasets are provided in the appendix.

We adopt the same evaluation metrics with previous works \cite{lei2021qvhighlights,lin2023univtg,liu2022umt}. To be specific, we compute Recall@1 with IoU threshold $\theta_{IoU}=0.5$ and $0.7$, mean average precision (mAP) with $\theta_{IoU}=0.5$ and $0.7$, and mAP with a series of thresholds [$0.5$:$0.05$:$0.95$] for MR on QVHighlights \cite{lei2021qvhighlights}. mAP and HIT@1 where positive samples are defined as with the saliency score of \textit{Very Good} are adopted for HD. On Ego4D-NLQ \cite{grauman2022ego4d}, Charades-STA \cite{gao2017tall}, and TACoS \cite{regneri2013grounding} datasets, we utilize Recall@1 with $\theta_{IoU}=\{0.3, 0.5, 0.7\}$ and mIoU to measure the MR performance. On YouTube Highlights \cite{sun2014ranking} for HD and TVSum \cite{song2015tvsum} for VS, we follow \cite{liu2022umt,lin2023univtg} and use the same \texttt{train/val} split with mAP and Top-5 mAP as metrics, respectively.

\begin{table}[t]
\setlength{\tabcolsep}{1.25pt}
\fontsize{6.5pt}{7.5pt}\selectfont
\caption{Class-wise video highlight detection results (mAP) on YouTube Highlights dataset. \hspace{-0.5mm}\protect\clipsdot\hspace{0.5mm}, \hspace{-0.5mm}\protect\sfdot\hspace{0.5mm}, and \hspace{-0.5mm}\protect\panndot\hspace{0.5mm} denote image, video, and audio backbones, respectively.}
\label{tab:youtube}
\centering
\begin{tabularx}{0.7\linewidth}{@{\hspace{1mm}}p{2.2cm}|@{\hspace{0.5mm}}p{0.8cm}<{\centering}p{0.8cm}<{\centering}p{0.8cm}<{\centering}p{0.8cm}<{\centering}p{0.8cm}<{\centering}p{0.8cm}<{\centering}p{0.8cm}<{\centering}}
\toprule
\textbf{Method} & \textbf{Dog} & \textbf{Gym.} & \textbf{Par.} & \textbf{Ska.} & \textbf{Ski.} & \textbf{Sur.} & \textbf{Avg.} \\
\midrule
\hspace{-1mm}\sfdot\hspace{0.1mm} RRAE \cite{yang2015unsupervised} & 49.0 & 35.0 & 50.0 & 25.0 & 22.0 & 49.0 & 38.3 \\
\hspace{-1mm}\sfdot\hspace{0.1mm} GIFs \cite{gygli2016video2gif} & 30.8 & 33.5 & 54.0 & 55.4 & 32.8 & 54.1 & 46.4 \\
\hspace{-1mm}\sfdot\hspace{0.1mm} LSVM \cite{sun2014ranking} & 60.0 & 41.0 & 61.0 & 62.0 & 36.0 & 61.0 & 53.6 \\
\hspace{-1mm}\sfdot\hspace{0.1mm} LIM-S \cite{xiong2019less} & 57.9 & 41.7 & 67.0 & 57.8 & 48.6 & 65.1 & 56.4 \\
\hspace{-1mm}\sfdot\hspace{0.1mm} SL-Module \cite{xu2021cross} & 70.8 & 53.2 & 77.2 & 72.5 & 66.1 &76.2 & 69.3 \\
\hspace{-1mm}\sfdot\hspace{0.1mm} PLD \cite{wei2022learning} & \underline{74.9} & 70.2 & 77.9 & 57.5 & 70.7 & 79.0 & 73.0 \\
\midrule
\hspace{-1mm}\vadot\hspace{0.1mm} MINI-Net \cite{hong2020mini} & 58.2 & 61.7 & 70.2 & 72.2 & 58.7 & 65.1 & 64.4 \\
\hspace{-1mm}\vadot\hspace{0.1mm} TCG \cite{ye2021temporal} & 55.4 & 62.7 & 70.9 & 69.1 & 60.1 & 59.8 & 63.0 \\
\hspace{-1mm}\vadot\hspace{0.1mm} Joint-VA \cite{badamdorj2021joint} & 64.5 & 71.9 & 80.8 & 62.0 & \underline{73.2} & 78.3 & 71.8 \\
\hspace{-1mm}\vadot\hspace{0.1mm} CO-AV \cite{li2022probing} & 60.9 & 66.0 & \textbf{89.0} & 74.1 & 69.0 & 81.1 & 74.7 \\
\hspace{-1mm}\vadot\hspace{0.1mm} UMT \cite{liu2022umt} & 65.9 & \underline{75.2} & \underline{81.6} & 71.8 & 72.3 & \underline{82.7} & 74.9 \\
\hspace{-1mm}\clipsfdot\hspace{0.1mm} UniVTG \cite{lin2023univtg} & 71.8 & \textbf{76.5} & 73.9 & \underline{73.3} & \underline{73.2} & 82.2 & \underline{75.2} \\
\midrule
\hspace{-1mm}\clipsdot\hspace{0.1mm} \modelbold{} & \textbf{75.6} & 73.5 & 73.0 & \textbf{74.6} & \textbf{74.8} & \textbf{84.8} & \textbf{76.1} \\
\bottomrule
\end{tabularx}
\end{table}

\begin{table}[t]
\setlength{\tabcolsep}{1.25pt}
\fontsize{6.5pt}{7.5pt}\selectfont
\caption{Class-wise video summarization results (Top-5 mAP) on TVSum dataset. \hspace{-0.5mm}\protect\clipsdot\hspace{0.5mm}, \hspace{-0.5mm}\protect\sfdot\hspace{0.5mm}, and \hspace{-0.5mm}\protect\panndot\hspace{0.5mm} denote image, video, and audio backbones, respectively. Our method does not require any additional video/audio features.}
\label{tab:tvsum}
\centering
\begin{tabularx}{0.825\linewidth}{@{\hspace{1mm}}p{2.2cm}|@{\hspace{1.5mm}}p{0.6cm}<{\centering}p{0.6cm}<{\centering}p{0.6cm}<{\centering}p{0.6cm}<{\centering}p{0.6cm}<{\centering}p{0.6cm}<{\centering}p{0.6cm}<{\centering}p{0.6cm}<{\centering}p{0.6cm}<{\centering}p{0.6cm}<{\centering}p{0.6cm}<{\centering}}
\toprule
\textbf{Method} & \textbf{VT} & \textbf{VU} & \textbf{GA} & \textbf{MS} & \textbf{PK} & \textbf{PR} & \textbf{FM} & \textbf{BK} & \textbf{BT} & \textbf{DS} & \textbf{Avg.} \\
\midrule
\hspace{-1mm}\sfdot\hspace{0.1mm} sLSTM \cite{zhang2016video} & 41.1 & 46.2 & 46.3 & 47.7 & 44.8 & 46.1 & 45.2 & 40.6 & 47.1 & 45.5 & 45.1 \\
\hspace{-1mm}\sfdot\hspace{0.1mm} SG \cite{mahasseni2017unsupervised} & 42.3 & 47.2 & 47.5 & 48.9 & 45.6 & 47.3 & 46.4 & 41.7 & 48.3 & 46.6 & 46.2 \\
\hspace{-1mm}\sfdot\hspace{0.1mm} LIM-S \cite{xiong2019less} & 55.9 & 42.9 & 61.2 & 54.0 & 60.4 & 47.5 & 43.2 & 66.3 & 69.1 & 62.6 & 56.3 \\
\hspace{-1mm}\sfdot\hspace{0.1mm} Trailer \cite{wang2020trailer} & 61.3 & 54.6 & 65.7 & 60.8 & 59.1 & 70.1 & 58.2 & 64.7 & 65.6 & 68.1 & 62.8 \\
\hspace{-1mm}\sfdot\hspace{0.1mm} SL-Module \cite{xu2021cross} & 86.5 & 68.7 & 74.9 & \textbf{86.2} & 79.0 & 63.2 & 58.9 & 72.6 & 78.9 & 64.0 & 73.3 \\
\hspace{-1mm}\sfdot\hspace{0.1mm} PLD \cite{wei2022learning} & 84.5 & 80.9 & 70.3 & 72.5 & 76.4 & \underline{87.2} & 71.9 & 74.0 & 74.4 & \underline{79.1} & 77.1 \\
\midrule
\hspace{-1mm}\vadot\hspace{0.1mm} MINI-Net \cite{hong2020mini} & 80.6 & 68.3 & 78.2 & 81.8 & 78.1 & 65.8 & 57.8 & 75.0 & 80.2 & 65.5 & 73.2 \\
\hspace{-1mm}\vadot\hspace{0.1mm} TCG \cite{ye2021temporal} & 85.0 & 71.4 & 81.9 & 78.6 & 80.2 & 75.5 & 71.6 & 77.3 & 78.6 & 68.1 & 76.8 \\
\hspace{-1mm}\vadot\hspace{0.1mm} Joint-VA \cite{badamdorj2021joint} & 83.7 & 57.3 & 78.5 & \underline{86.1} & 80.1 & 69.2 & 70.0 & 73.0 & \textbf{97.4} & 67.5 & 76.3 \\
\hspace{-1mm}\vadot\hspace{0.1mm} CO-AV \cite{li2022probing} & \textbf{90.8} & 72.8 & 84.6 & 85.0 & 78.3 & 78.0 & 72.8 & 77.1 & 89.5 & 72.3 & 80.1 \\
\hspace{-1mm}\vadot\hspace{0.1mm} UMT \cite{liu2022umt} & \underline{87.5} & 81.5 & 88.2 & 78.8 & 81.4 & 87.0 & \underline{76.0} & 86.9 & 84.4 & \textbf{79.6} & \underline{83.1} \\
\hspace{-1mm}\clipsfdot\hspace{0.1mm} UniVTG \cite{lin2023univtg} & 83.9 & \underline{85.1} & \underline{89.0} & 80.1 & \underline{84.6} & 81.4 & 70.9 & \textbf{91.7} & 73.5 & 69.3 & 81.0 \\
\midrule
\hspace{-1mm}\clipsdot\hspace{0.1mm} \modelbold{} & 85.0 & \textbf{85.9} & \textbf{91.0} & 81.7 & \textbf{88.8} & \textbf{87.4} & \textbf{78.1} & \underline{89.2} & \underline{90.3} & 74.7 & \textbf{85.2} \\
\bottomrule
\end{tabularx}
\end{table}

\subsection{Implementation Details}

In all experiments, we adopt the visual and text encoders of CLIP ViT-B/32 \cite{radford2021learning} as our backbones. The whole CLIP model is frozen during training, \ie, only the parameters in \block{}, temporal feature pyramid, prediction heads are learnable. Without further specification, we set $K = 4$, \ie, attaching \block{} to the last 4 layers of CLIP encoders. In \block, the hidden size is set to $256$. Only one transformer layer with \textit{post-norm} style and 8 attention heads is used for temporal modeling. The number of layers for the temporal feature pyramid is set to $4$ for QVHighlights, Ego4D-NLQ, Charades-STA, and TACoS, while $1$ for other datasets. Please refer to the appendix for details.

\begin{figure}[t]
\centering
\begin{minipage}{\linewidth}
\includegraphics[width=\textwidth]{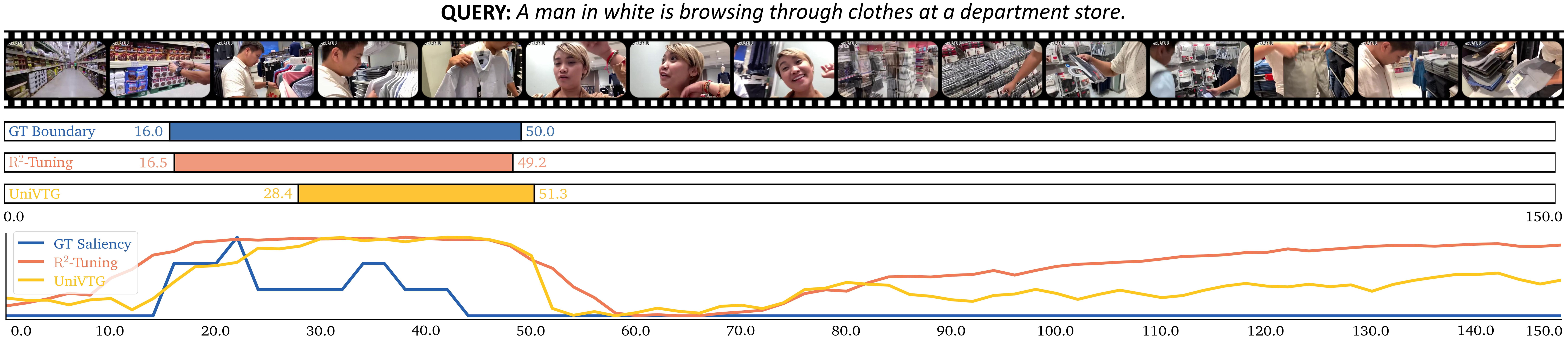}
\vspace{0.1mm}
\end{minipage}
\begin{minipage}{\linewidth}
\includegraphics[width=\textwidth]{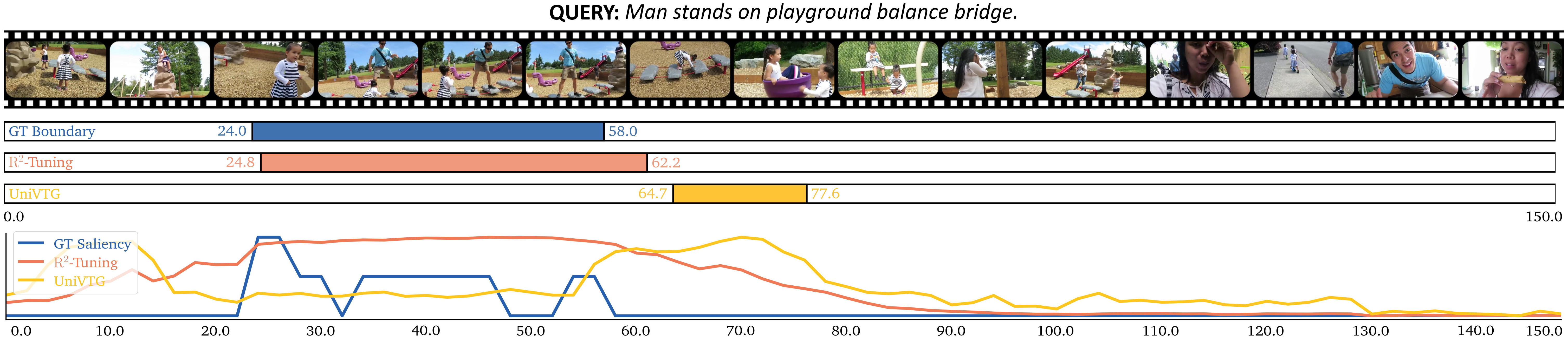}
\end{minipage}
\caption{Visualization of joint moment retrieval and highlight detection results on QVHighlights \texttt{val} split. We compare our method with UniVTG \cite{lin2023univtg}, which is a strong baseline leveraging both additional temporal backbone and large-scale post pre-training. \model{} can accurately regress the boundaries of moments and predict highlight saliency scores through its novel feature refinement design.}
\label{fig:vis}
\end{figure}

\subsection{Comparison with State-of-the-Arts}

\paragraphbf{Joint Moment Retrieval and Highlight Detection}

We first evaluate our method on QVHighlights \cite{lei2021qvhighlights} \texttt{test} split, the only dataset that supports both moment retrieval and highlight detection. The results are shown in Table~\ref{tab:qvhighlights}. The first group of methods utilizes more than one feature extractor. The second group is that with large-scale post pre-training (defined as the extra pre-training stage between loading backbone weights and training on downstream datasets.). Without any additional backbone and extra pre-training, \model{} achieves start-of-the-art performance compared with all previous methods with the fewest learnable parameters. Figure~\ref{fig:vis} visualizes the model predictions.

\paragraphbf{Moment Retrieval}

We then evaluate our model on moment retrieval task in egocentric \cite{grauman2022ego4d} and in-door \cite{regneri2013grounding,gao2017tall} domains. The results are shown in Table~\ref{tab:moment}. We follow \cite{lin2023univtg} and compare only with the methods using CLIP + SlowFast features. \model{} still works better than all baselines with extra features. We also observe that \model{} shows its significance on high-quality retrievals (R1@0.7), which requires accurate temporal modeling. This benefits from the fine-grained temporal modeling ability of reversed and recurrent designs.

\paragraphbf{Highlight Detection \& Video Summarization}

The performances of highlight detection on YouTube Highlights \cite{sun2014ranking} and extractive video summarization on TVSum \cite{song2015tvsum} are reported in Table~\ref{tab:youtube} and Table~\ref{tab:tvsum}, respectively. Following previous works \cite{liu2022umt,lin2023univtg}, we trained the model separately on each domain. The first group of methods is based on video backbones, which naturally have basic temporal modeling abilities. The second group of methods is enhanced by extra features such as image and audio. \model{} can still reach the best performance when trained on small-scale data.

\subsection{Detailed Analysis}

\paragraphbf{Multi-Granularity Features}

\begin{figure}[t]
\centering
\includegraphics[width=\linewidth]{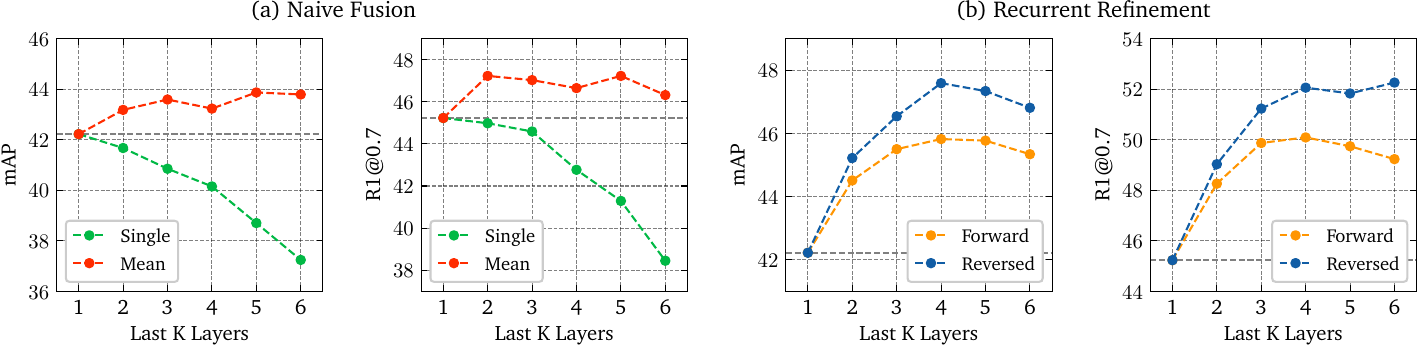}
\caption{Different feature refinement strategies for MR on QVHighlights \texttt{val} split. \textit{Single} means utilizing the $K$-th layer only, while \textit{Mean} denotes averaging the features from last $K$ layers. \textit{Forward} and \textit{Reversed} indicate different feature refinement directions.}
\label{fig:refine}
\end{figure}

To investigate the significance of granularity flexibility, we set up a simple baseline that only attaches \block{} to one layer of CLIP encoders. Then, we make use of the multi-layer information by averaging the features from the last $K$ layers. These two variants are compared in Figure~\ref{fig:refine}~(a). It can be observed that for the single-layer setting, higher-level features are more discriminative than lower-level ones. However, even a naive fusion strategy for the multi-layer information brings significant performance gains.

\begin{figure}[t]
\begin{minipage}{0.58\textwidth}
\centering
\includegraphics[height=4cm]{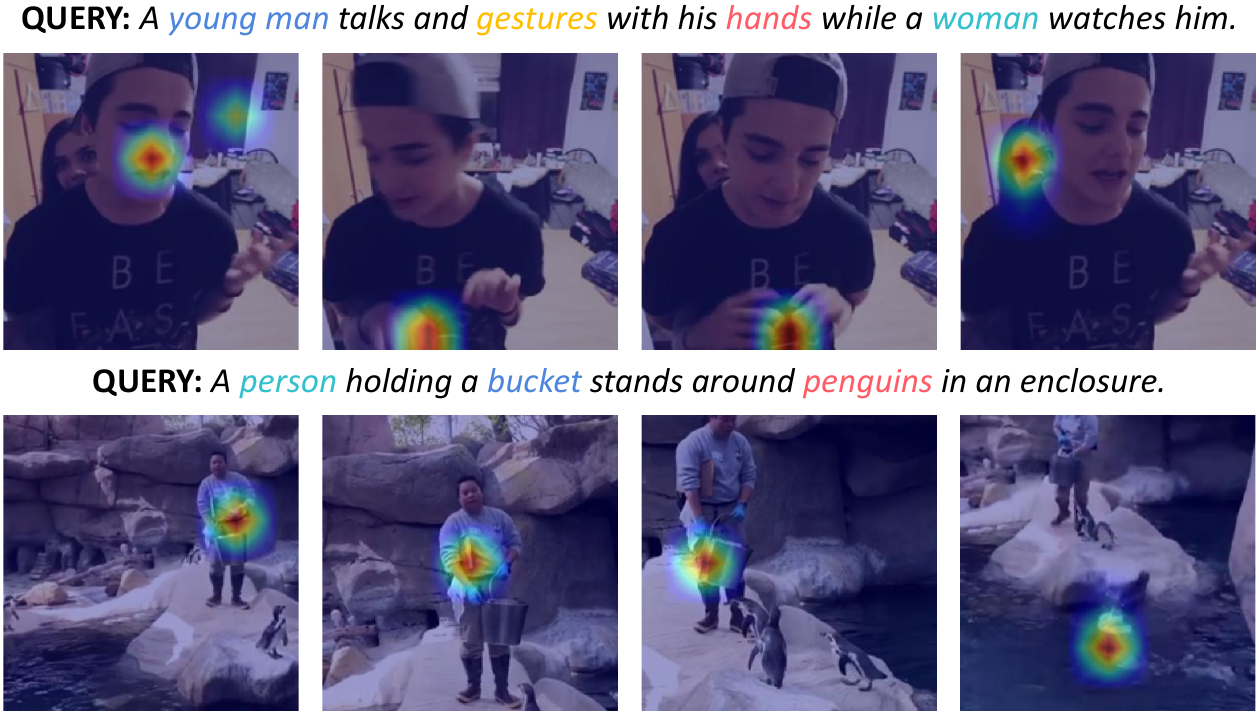}
\caption{Visualization of the attention maps for query-modulated spatial pooling. Diverse regions are learned for different queries.}
\label{fig:attention}
\end{minipage}
\hfill
\begin{minipage}{0.395\textwidth}
\centering
\includegraphics[height=4cm]{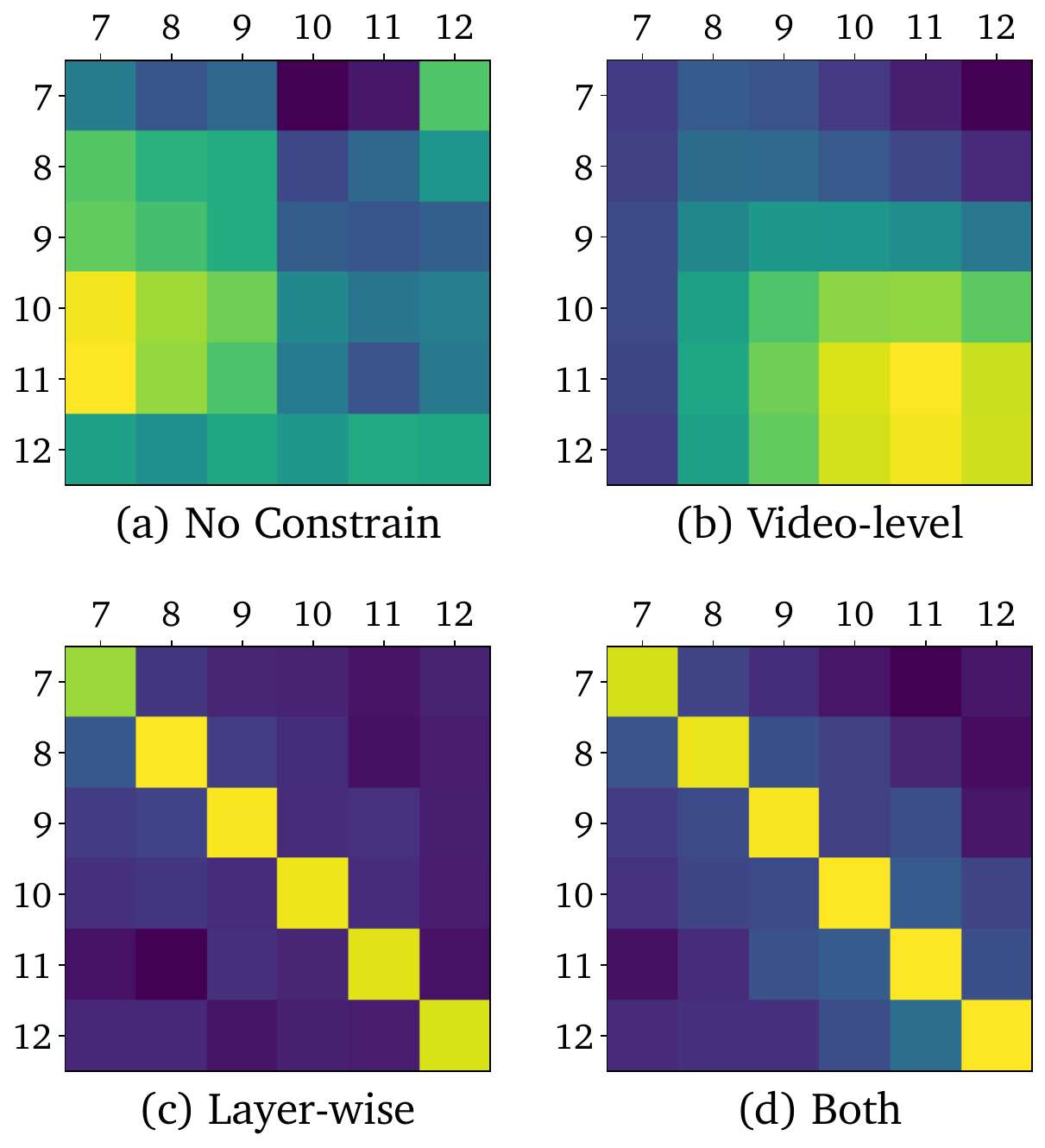}
\caption{Cosine similarities of video (x-axis) and text (y-axis) features among 7$\sim$12 CLIP layers.}
\label{fig:calibration}
\end{minipage}
\end{figure}

\paragraphbf{Reversed Recurrent Feature Refinement}

We then further justify our design by comparing two refinement directions (\ie, forward and reversed) in Figure~\ref{fig:refine}~(b). Note that when $K\hspace{-1mm}=\hspace{-1mm}1$, all four designs in Figure~\ref{fig:refine} are of the same architecture. When utilizing more layers, both forward and reversed refinement benefit from more information, while the reversed scheme statically performs better, as the multi-granularity features shall be refined from coarse to fine.

\paragraphbf{Query-Modulated Spatial Pooling}

We visualize the attention maps learned by query-modulated spatial pooling in Figure~\ref{fig:attention}. A query generally has multiple keywords that provide discriminative information to localize moments. The token-wise attention in Eq.~\ref{eq:attention} can guide the model to focus on multiple regions simultaneously, as can be seen in the diverse but meaningful patterns.

\begin{table}[t]
\setlength{\tabcolsep}{1.25pt}
\fontsize{6.5pt}{7.5pt}\selectfont
\centering
\begin{minipage}{0.515\textwidth}
\caption{Effectiveness justification of granularity calibration on QVHighlights \texttt{val} split.}
\label{tab:calibration}
\begin{tabularx}{\linewidth}{@{\hspace{1mm}}cccccccccc}
\toprule
\multirow{2.5}{*}{$\mathcal{L}_{video}$} & \multirow{2.5}{*}{$\mathcal{L}_{layer}$} && \multicolumn{3}{c}{\textbf{MR}} && \multicolumn{2}{c}{\textbf{HD}} \\
\cmidrule{4-6}\cmidrule{8-9}
&&& R1@0.5 & R1@0.7 & mAP && mAP & HIT@1 \\
\midrule
&&& 64.48 & 48.60 & 44.01 && 37.94 & 62.67 \\
\cmark &&& \underline{67.68} & \underline{51.61} & \underline{46.74} && \underline{39.81} & \underline{65.16} \\
& \cmark && 64.71 & 48.84 & 44.60 && 38.91 & 63.35 \\
\cmark & \cmark && \textbf{68.71} & \textbf{52.06} & \textbf{47.59} && \textbf{40.59} & \textbf{64.32} \\
\bottomrule
\end{tabularx}
\end{minipage}
\hfill
\begin{minipage}{0.46\textwidth}
\caption{MR mAP with different query lengths on QVHighlights \texttt{val} split.}
\label{tab:length}
\begin{tabularx}{\linewidth}{@{\hspace{0.5mm}}p{1.7cm}<{\raggedright}ccccc}
\toprule
\multirow{2.65}{*}{\hspace{3mm}\textbf{Method}} & \multicolumn{5}{c}{\textbf{\#Words}} \\
\cmidrule{2-6}
& \textbf{1-10} & \textbf{11-20} & \textbf{21-30} & \textbf{31-40} & \textbf{$\geqslant$41} \\
\midrule
QD-DETR \cite{moon2023query} & \underline{40.30} & \underline{42.32} & \textbf{29.01} & 0.10 & 26.67 \\
EaTR \cite{jang2023knowing} & 38.94 & 39.54 & 24.58 & \underline{12.03} & \underline{41.43} \\
UniVTG \cite{lin2023univtg} & 32.38 & 32.65 & 20.88 & 0.00 & 31.11 \\
\modelbold & \textbf{47.08} & \textbf{49.13} & \underline{28.79} & \textbf{67.24} & \textbf{72.38} \\
\bottomrule
\end{tabularx}
\end{minipage}
\end{table}

\paragraphbf{Granularity Calibration}

To verify the effectiveness of granularity calibration, we also visualize the cosine distances of all the visual-query pairs from 7\,$\sim$\,12 layers in Fig.~\ref{fig:calibration}. Before calibration, the visual-text features are not well-aligned. Adding the video-level constraint makes higher-level features aligned with each other (b), while adding layer-wise constraint makes the model distill diverse information across different layers (c). The diagonal in (d) verifies that both calibration constraints can maximize the mutual similarities of paired visual and text features. Some quantitative results are presented in Table~\ref{tab:calibration}. Both video-level and layer-wise contrastive can help align the multi-granularity visual \& query semantics, while their combination can further boost the performances. We also group the queries in QVHighlights into 5 bins with different lengths (acting as multi-granularities) and compute the means of MR mAP in Tab.~\ref{tab:length}. Although most training queries are coarse-grained ($\leqslant$ 30 words), \model{} can easily generalize to fine-grained queries ($\geqslant$ 31 words), surpassing previous methods.

\section{Conclusion}

This paper introduces \model, a parameter- and memory-efficient transfer learning framework for video temporal grounding. It learns a lightweight side-adapter (\block) that is recurrently attached to the last few layers of a frozen CLIP to adaptively pool spatial details and refine temporal correlations. Two contrastive constraints (video-level and layer-wise) are utilized to calibrate the granularities of CLIP visual and text encoders. Experiments across three VTG tasks on six public benchmarks demonstrate the significance and effectiveness of the proposed scheme. We hope that the proposed framework sparks further research on efficient image-to-video transfer learning for untrimmed videos.

\section*{Acknowledgments}

This work was supported in part by Hong Kong Research Grants Council GRF-15229423, US NIH grant R01HD104969, and NSF award IIS-2239688.

\appendix

\section*{Appendix}

In the appendix, we provide more descriptions of the model, datasets, and implementation details to complement the main paper. Additional experiments, visualizations, and discussions on limitations \& future work are also incorporated to justify the effectiveness of our method.

\section{Preliminary: The CLIP Model}

The philosophy of CLIP \cite{radford2021learning} is to regard images and their corresponding descriptions as different views of the same concept, and to align them in the semantic space. The CLIP model has a two-stream architecture, consisting of a visual encoder (\eg, ResNet \cite{he2016deep}, ViT \cite{dosovitskiy2020image}) and a text encoder (\eg, Transformer \cite{vaswani2017attention}). In this work, we only consider the case of using ViT as visual encoder, since it shares a unified architecture with text encoder and has better performance on representation learning. The input image $I$ is first split into non-overlapping patches, while the text $T$ is also tokenized into discrete tokens. They are then mapped into a shared semantic space by the two encoders. This process is optimized using a symmetric InfoNCE \cite{oord2018representation} loss $\mathcal{L}$:
\begin{gather}
\mathcal{L}_{I \to T} = -\frac{1}{B}\sum_{i=1}^B\log\frac{\exp(\mathbf{v}_i^\top\mathbf{q}_i/\tau)}{\sum_{j=1}^B \exp(\mathbf{v}_i^\top\mathbf{q}_j/\tau)} \\
\mathcal{L}_{T \to I} = -\frac{1}{B}\sum_{i=1}^B\log\frac{\exp(\mathbf{q}_i^\top\mathbf{v}_i/\tau)}{\sum_{j=1}^B \exp(\mathbf{q}_i^\top\mathbf{v}_j/\tau)} \\
\mathcal{L} = \frac{1}{2}(\mathcal{L}_{I \to T} + \mathcal{L}_{T \to I})
\end{gather}
Here, $\mathbf{v}$ and $\mathbf{q}$ are encoded image and text embeddings, $i$ and $j$ are indices of samples in a batch with size $B$, and $\tau$ is a temperature parameter controlling the smoothness of the softmax distribution. After training, the paired $\mathbf{v}$ and $\mathbf{q}$ shall be well-aligned in their shared semantic space.

\section{Detailed Comparison with Counterparts}\label{sec:design}

In order to further justify our novelty and distinguish our design from previous methods, we provide an in-depth comparison between our method and existing ones. To evaluate the effectiveness of a transfer learning method, we define four design principles leading to a good transfer learning framework for VTG, explained as follows:

\vspace{5mm}
\begin{minipage}{11.65cm}
\begin{enumerate}[label=\textbf{P\arabic*.}]
\item \textbf{Strong Spatial-Temporal Modeling Ability:} VTG requires the model to regress moment temporal boundaries or densely predict frame-level scores/indicators. Hence, strong spatial-temporal modeling abilities are needed to accurately ground the video.
\end{enumerate}
\end{minipage}
\vspace{5mm}

\vspace{5mm}
\begin{minipage}{11.65cm}
\begin{enumerate}[label=\textbf{P\arabic*.},start=2]
\item \textbf{Parameter-Efficient:} As tuning all the parameters of CLIP is resource-consuming and may even lead to catastrophic forgetting \cite{kirkpatrick2017overcoming}, the transfer learning framework should only learn a few parameters to take full advantage of the pre-trained knowledge.
\item \textbf{Memory-Efficient:} Although some transfer learning frameworks are parameter-efficient \cite{jia2022visual,pan2022st}, they might not be memory-efficient as computing the gradients of these learnable parameters requires back-propagating through the whole network, consuming a large amount of GPU memory.
\item \textbf{Granularity-Flexible:} As mentioned in the main paper, queries for VTG might be of different granularity levels. The capability of parsing multi-granularity queries and extracting apprapriate features accordingly is essential for flexible VTG.
\end{enumerate}
\end{minipage}
\vspace{5mm}

\begin{table}[t]
\setlength{\tabcolsep}{5pt}
\fontsize{8.5pt}{10.5pt}\selectfont
\caption{Capability comparison among different image-to-video transfer learning frameworks. The design principles P1$\hspace{0.5mm}\sim\hspace{0.5mm}$P4 are defined in Section~\ref{sec:design}. Our proposed \model{} is the only one satisfying all the principles.}
\label{tab:capability}
\centering
\begin{tabularx}{0.735\linewidth}{p{5.5cm}<{\raggedright}|cccc}
\toprule
\textbf{Framework} & \textbf{P1} & \textbf{P2} & \textbf{P3} & \textbf{P4} \\
\midrule
(a) Post-processing \cite{lei2021qvhighlights,liu2022umt,moon2023query,jang2023knowing} & \rmark & \gmark & \gmark & \rmark \\
(b) Full/Partial Tuning \cite{rasheed2023fine,yan2023unloc} & \omark & \rmark & \rmark & \rmark \\
(c) Adapter/Prompt Tuning \cite{jia2022visual,pan2022st} & \gmark & \gmark & \rmark & \rmark \\
\midrule
(d) \modelbold{} (Ours) & \gmark & \gmark & \gmark & \gmark \\
\bottomrule
\end{tabularx}
\end{table}

We then systematically compare our proposed scheme with previous ones from two aspects, \textit{architecture} and \textit{capabilities}, shown in Figure~{\color[HTML]{FF0000}3}~(main paper) and Table~\ref{tab:capability}, respectively.

\paragraphbf{Architecture}

As illustrated in Figure~{\color[HTML]{FF0000}3}~(main paper), \textbf{Post-processing} methods \cite{lei2021qvhighlights,liu2022umt,xu2023mh,moon2023query,jang2023knowing,luo2022clip4clip} follow a straightforward strategy that simply regarding CLIP as a spatial \& textual feature extractor. All the temporal modeling capabilities are realized by carefully designed modules attached to the final outputs of CLIP. An auxiliary temporal backbone (\eg, SlowFast \cite{feichtenhofer2019slowfast}) is also utilized to complement the spatial-temporal information. \textbf{Full/Partial Tuning} methods \cite{rasheed2023fine,yan2023unloc} unfreeze either the visual or textual stream of CLIP, and jointly learn it with temporal modeling modules. These models are usually combined with post pre-training due to the potential catastrophic forgetting issue \cite{kirkpatrick2017overcoming}. \textbf{Adapter/Prompt Tuning} methods \cite{jia2022visual,pan2022st,ju2022prompting,bain2022clip,rasheed2023fine,huang2023vop} learn lightweight adapters or continuous prompt tokens to transfer the pre-trained knowledge. These learnable sub-modules or tokens may perturb the original attention patterns \cite{petrov2023prompting}. Our \modelbold, instead, learns an adapter that be used as a side module \cite{sung2022lst,qing2023disentangling,diao2023unipt} for the main network, preserving the information patterns in the main model while taking advantages of the multi-level representations. Compared with the recent side-tuning/parallel-tuning methods, our approach shows its advantages by leveraging the \textit{reversed} fusion path and the \textit{recurrent} refinement strategy.

\paragraphbf{Capabilities}

Table~\ref{tab:capability} compares the capabilities of different image-to-video transfer learning frameworks. \textbf{Post-processing} methods \cite{lei2021qvhighlights,liu2022umt,xu2023mh,moon2023query,jang2023knowing,luo2022clip4clip} are naturally parameter- and memory-efficient, as they only require to learn temporal modules on top of pre-extracted CLIP features. Nevertheless, these methods largely rely on the auxiliary temporal backbone as frame-level CLIP embeddings limit the temporal modeling capabilities. Merely using the last hidden features is not granularity-flexible as well. \textbf{Full/Partial Tuning} methods \cite{rasheed2023fine,yan2023unloc} may have slightly better spatial-temporal modeling as they fine-tuned the spatial encoders on videos. But the large amount of learnable parameters make the fine-tuning process inefficient. \textbf{Adapter/Prompt Tuning} methods \cite{jia2022visual,pan2022st,ju2022prompting,bain2022clip,rasheed2023fine,huang2023vop} could model spatial-temporal information by designing special adapters \cite{pan2022st} or prompts \cite{huang2023vop}. However, training these adapters/prompts might consume high GPU memory, as the learnable parameters are tightly bound with frozen ones, thus the gradient computation cannot be disentangled. To our best knowledge, \modelbold{} is the only scheme that satisfies all the design principles.

\section{Datasets}

Following the previous work \cite{lin2023univtg}, we conduct experiments on three VTG tasks across six datasets, which are introduced in detail as follows.

\paragraphbf{Joint Moment Retrieval and Highlight Detection}

We adopt QVHighlights \cite{lei2021qvhighlights} since this is the only dataset that supports both moment retrieval and highlight detection. It contains $10,148$ videos, each trimmed to $\sim\hspace{-0.5mm}150$s long. The domains of the videos cover from daily vlogs, travel vlogs, to news events. A total of $10,310$ queries with $18,367$ disjoint moments are manually annotated. This dataset also has a private \texttt{test} split, so that the evaluation on this dataset shall be more reliable than on others.

\paragraphbf{Moment Retrieval}

As for moment retrieval, we adopt Ego4D-NLQ (Natural Language Query) \cite{grauman2022ego4d}, Charades-STA \cite{gao2017tall}, and TACoS \cite{regneri2013grounding} datasets. Ego4D-NLQ contains $1.3$K videos with $8\hspace{-0.5mm}\sim\hspace{-0.5mm}20$ minutes durations under daily egocentric scenarios. $15.2$K queries in the form of questions (\eg, \textit{where did I put the plier}) are annotated with precise moments. Charades-STA is an extension of the Charades \cite{sigurdsson2016hollywood} dataset with moment annotations. It has $9,848$ unique in-door videos with an average length of $30.6$s, and a total of $16,128$ query-moment pairs. TACoS is a small-scale dataset with $273$ videos, each of $\sim\hspace{-0.5mm}4.8$ minutes long. There are $9,790$, $4,436$, and $4,001$ queries for training, validation, and testing, respectively.

\paragraphbf{Highlight Detection}

We utilize YouTube Highlights \cite{sun2014ranking} dataset to evaluate the highlight detection performance. As some of the videos are no longer available, we follow previous works \cite{liu2022umt,lin2023univtg} to use a subset that contains $435$ videos under $6$ domains. The video domains are used as queries.

\paragraphbf{Video Summarization}

TVSum \cite{song2015tvsum} dataset is leveraged for video summarization evaluation. It contains $10$ domains, each with $5$ videos. We follow the same setting as previous works \cite{liu2022umt,lin2023univtg} that use $4$ videos for training and $1$ video for testing. The video titles are used as queries. Note that TVSum \cite{song2015tvsum} is regarded as a highlight detection dataset in \cite{lin2023univtg} but as a video summarization dataset in our work. We claim that both choices are reasonable as HD and VS essentially share similar problem formulations. The only differences are that: 1) only a small proportion (\eg, 20\%) of the video clips are regarded as summary for VS, while there is no limitation on number of highlight clips for HD, and 2) VS requires to quantize the output scores to binary values in practice. Previous works on HD \cite{hong2020mini,xu2021cross,badamdorj2021joint,liu2022umt} and VS \cite{gygli2014creating,apostolidis2021video,jiang2022joint} also mixed these datasets when conducting evaluations.

\begin{table}[t]
\setlength{\tabcolsep}{3pt}
\fontsize{8.5pt}{10.5pt}\selectfont
\caption{Hyperparameters used for different datasets. FPS means frame rate, LR denotes learning rate, Epochs denotes total training epochs, Warmup means number of warmup iterations, and LR Drop means the epoch that drops learning rate by $1/10$. * Video clips in YouTube Highlights have overlaps. We use the same setting as \cite{liu2022umt}.}
\label{tab:implementation}
\centering
\begin{tabularx}{\linewidth}{p{3.6cm}<{\raggedright}|@{\hspace{3mm}}cccccc}
\toprule
\textbf{Dataset} & \textbf{FPS} & \textbf{Batch Size} & \textbf{LR} & \textbf{Epochs} & \textbf{Warmup} & \textbf{LR Drop} \\
\midrule
QVHighlights \cite{lei2021qvhighlights} & $0.5$ & $128$ & $5e^{-4}$ & $30$ & $500$ & $20$ \\
\midrule
Ego4D-NLQ \cite{grauman2022ego4d} & $0.5$ & $32$ & $2.5e^{-4}$ & $30$ & $500$ & $20$ \\
Charades-STA \cite{gao2017tall} & $1.0$ & $32$ & $2.5e^{-4}$ & $50$ & $500$ & $30$ \\
TACoS \cite{regneri2013grounding} & $0.5$ & $32$ & $2.5e^{-4}$ & $100$ & $500$ & $50$ \\
\midrule
YouTube Highlights \cite{sun2014ranking} & * & $4$ & $5e^{-4}$ & $200$ & $50$ & -- \\
TVSum \cite{song2015tvsum} & $0.5$ & $4$ & $5e^{-4}$ & $500$ & $50$ & -- \\
\bottomrule
\end{tabularx}
\end{table}

\section{Implementation Details}

Table~\ref{tab:implementation} lists the hyperparameters used for different datasets. We set the strides of temporal feature pyramid to $(1, 2, 4, 8)$ by default. For all videos, we directly resize and crop the frames to $224 \times 224$ without any augmentations. The same regularization methods with CLIP \cite{radford2021learning} pre-training is adopted to aligned the inputs. The maximum number of query tokens is set to $77$. We train our model on a single NVIDIA A100 (80G) GPU. Automatic mixed precision (AMP) with FP16 is utilized to accelerate training.

\section{More Experiments}

We conduct further experiments and ablation studies on QVHighlights \cite{lei2021qvhighlights} and Ego4D-NLQ \cite{grauman2022ego4d} datasets to justify the effectiveness of our method. These two datasets are selected because 1) QVHighlights is the only dataset supporting both MR and HD, and 2) Ego4D-NLQ has much longer videos and different domains/queries compared with QVHighlights.

\begin{table}[t]
\setlength{\tabcolsep}{3pt}
\fontsize{8.5pt}{10.5pt}\selectfont
\caption{Comparison among different PETL and METL methods on QVHighlights \texttt{val} split. \#Params and Memory represent the number of learnable parameters and peak GPU memory ($224 \times 224$ inputs with $32$ batch size \& FP16), respectively.}
\label{tab:petl}
\centering
\begin{tabularx}{0.97\linewidth}{@{\hspace{2mm}}p{2.1cm}<{\raggedright}p{1.5cm}<{\centering}p{1.5cm}<{\centering}ccccccc}
\toprule
\multirow{2.5}{*}{\hspace{5mm}\textbf{Method}} & \multirowcell{2.5}{\textbf{\#Params} \\ \textbf{(M)}} & \multirowcell{2.5}{\textbf{Memory} \\ \textbf{(GB)}} && \multicolumn{3}{c}{\textbf{MR}} && \multicolumn{2}{c}{\textbf{HD}} \\
\cmidrule{5-7}\cmidrule{9-10}
&&&& R1@0.5 & R1@0.7 & mAP && mAP & HIT@1 \\
\midrule
w/o Tuning & \underline{2.7} & \textbf{4.3} && 64.19 & 45.03 & 42.22 && 39.05 & 63.16 \\
\midrule
Full-Tuning & 147.0 & 58.5 && 64.69 & 45.86 & 42.41 && 38.55 & 62.50 \\
LoRA \cite{hu2021lora} & 4.4 & 55.4 && 63.70 & 44.93 & 41.29 && 38.47 & 61.09 \\
VPT \cite{jia2022visual} & 2.8 & 42.9 && 64.55 & 45.97 & 42.36 && 38.56 & 62.48 \\
VoP \cite{huang2023vop} & 14.5 & 45.2 && 65.23 & 46.01 & 42.45 && 38.70 & 62.75 \\
ST-Adapter \cite{pan2022st} & 12.2 & 44.3 && 65.19 & \underline{48.58} & 44.27 && 39.52 & 63.49 \\
E$^3$VA \cite{yin2023parameter} & 3.3 & 10.8 && 65.10 & 48.38 & \underline{45.06} && 39.45 & 63.38 \\
LoSA \cite{mercea2024time} & 4.8 & 10.9 && 65.06 & 48.43 & 44.97 && 39.38 & 63.35 \\
LST \cite{sung2022lst} & 7.4 & 10.8 && \underline{65.26} & 48.42 & 45.03 && \underline{39.81} & \textbf{64.39} \\
\midrule
\modelbold & \textbf{2.7} & \underline{10.6} && \textbf{68.71} & \textbf{52.06} & \textbf{47.59} && \textbf{40.59} & \underline{64.32} \\
\bottomrule
\end{tabularx}
\end{table}

\paragraphbf{Comparison with Other Transfer Learning Methods}

Table~\ref{tab:petl} compares \model{} with existing representative PETL (LoRA \cite{hu2021lora}, VPT \cite{jia2022visual}, VoP \cite{huang2023vop}, and ST-Adapter \cite{pan2022st}) and METL (E$^3$VA \cite{yin2023parameter}, LoSA \cite{mercea2024time}, LST \cite{sung2022lst}) methods on QVHighlights. ``w/o Tuning'' means the entire CLIP is frozen and without any adapters, \ie, only the temporal feature pyramid and prediction heads are trainable. The \textit{rank} values for LoRA \cite{hu2021lora}, E$^3$VA \cite{yin2023parameter}, and LoSA \cite{mercea2024time} are set to $16$. Our model consumes the least learnable parameters and GPU memory, while achieving the best MR and HD performances, demonstrating its parameter- and memory-efficiency.

\begin{table}[t]
\setlength{\tabcolsep}{3pt}
\fontsize{8.5pt}{10.5pt}\selectfont
\caption{Performance comparison with different orders of attention layers. Self and Cross represent self-attention and cross-attention, respectively.}
\label{tab:order}
\centering
\begin{tabularx}{0.775\linewidth}{@{\hspace{2mm}}p{1.9cm}<{\centering}ccccccccccc}
\toprule
\multirow{2.5}{*}{\textbf{Order}} && \multicolumn{2}{c}{\textbf{MR@QVHL}} && \multicolumn{2}{c}{\textbf{HD@QVHL}} && \multicolumn{2}{c}{\textbf{MR@Ego4D}} \\
\cmidrule{3-4}\cmidrule{6-7}\cmidrule{9-10}
&& R1@0.7 & mAP && mAP & HIT@1 && R1@0.7 & mIoU \\
\midrule
Self $\to$ Cross && 49.42 & 45.21 && 39.10 & 63.35 && 2.01 & 4.75 \\
Cross $\to$ Self && \textbf{52.06} & \textbf{47.59} && \textbf{40.59} & \textbf{64.32} && \textbf{2.12} & \textbf{4.94} \\
\bottomrule
\end{tabularx}
\end{table}

\paragraphbf{Order of Attention Layers}

Table~\ref{tab:order} compares different orders of attention layers. The standard transformer decoder \cite{vaswani2017attention} (Self $\to$ Cross) performs temporal modeling before aggregating query information, while the temporal modeling in our design (Cross $\to$ Self) can benefit from the guidance from query.

\paragraphbf{Effect of Recurrent Refinement}

Table~\ref{tab:recurrent} presents the justification of \textit{recurrent} design of \block. We compare it with a baseline that use a unique \block{} for each layer, \ie, parameters for different refinement steps are not shared. Even with much fewer parameters, adopting the \textit{recurrent} design does not lead to performance drops.

\begin{table}[t]
\setlength{\tabcolsep}{3pt}
\fontsize{8.5pt}{10.5pt}\selectfont
\caption{Effectiveness justification of the recurrent design (parameter sharing across layers) in \block.}
\label{tab:recurrent}
\centering
\begin{tabularx}{0.705\linewidth}{@{\hspace{2.5mm}}p{1cm}<{\centering}ccccccccccc}
\toprule
\multirow{2.5}{*}{\textbf{Shared}} && \multicolumn{2}{c}{\textbf{MR@QVHL}} && \multicolumn{2}{c}{\textbf{HD@QVHL}} && \multicolumn{2}{c}{\textbf{MR@Ego4D}} \\
\cmidrule{3-4}\cmidrule{6-7}\cmidrule{9-10}
&& R1@0.7 & mAP && mAP & HIT@1 && R1@0.7 & mIoU \\
\midrule
&& 51.35 & 47.48 && 40.28 & \textbf{65.42} && 2.06 & 4.83 \\
\cmark && \textbf{52.06} & \textbf{47.59} && \textbf{40.59} & 64.32 && \textbf{2.12} & \textbf{4.94} \\
\bottomrule
\end{tabularx}
\end{table}

\begin{table}[t]
\setlength{\tabcolsep}{3pt}
\fontsize{8.5pt}{10.5pt}\selectfont
\caption{Performance comparison with different numbers of transformer layers for temporal refinement.}
\label{tab:layers}
\centering
\begin{tabularx}{0.705\linewidth}{@{\hspace{2mm}}p{1.1cm}<{\centering}ccccccccccc}
\toprule
\multirow{2.5}{*}{\textbf{\#Layers}} && \multicolumn{2}{c}{\textbf{MR@QVHL}} && \multicolumn{2}{c}{\textbf{HD@QVHL}} && \multicolumn{2}{c}{\textbf{MR@Ego4D}} \\
\cmidrule{3-4}\cmidrule{6-7}\cmidrule{9-10}
&& R1@0.7 & mAP && mAP & HIT@1 && R1@0.7 & mIoU \\
\midrule
1 && 52.06 & 47.59 && 40.59 & 64.32 && 2.12 & 4.94 \\
2 && 52.14 & 47.62 && 40.63 & 64.28 && 2.15 & 4.97 \\
3 && \textbf{52.71} & 47.68 && 40.80 & 64.35 && \textbf{2.21} & \textbf{5.03} \\
4 && 52.32 & \textbf{47.81} && \textbf{40.85} & 63.30 && 2.19 & 4.91 \\
5 && 52.13 & 47.53 && 40.78 & \textbf{64.41} && 2.10 & 4.88 \\
6 && 51.16 & 47.42 && 40.61 & 64.33 && 2.04 & 4.82 \\
\bottomrule
\end{tabularx}
\end{table}

\paragraphbf{Number of Temporal Transformer Layers}

We study how the number of temporal transformer layers in \block{} affects the model performance. Table~\ref{tab:layers} shows the results. We observe that adding the number of transformer layers for temporal refinement only brings negligible performance gains. The best results appear when number of layers equals to $3$ or $4$. As discussed in the main paper, changing the number of refinement steps leads to much more significant reflections on performances, demonstrating the effectiveness of the \textit{recurrent} refinement design.

\section{More Visualizations}

\begin{figure*}
\centering
\begin{minipage}{\linewidth}
\includegraphics[width=\textwidth]{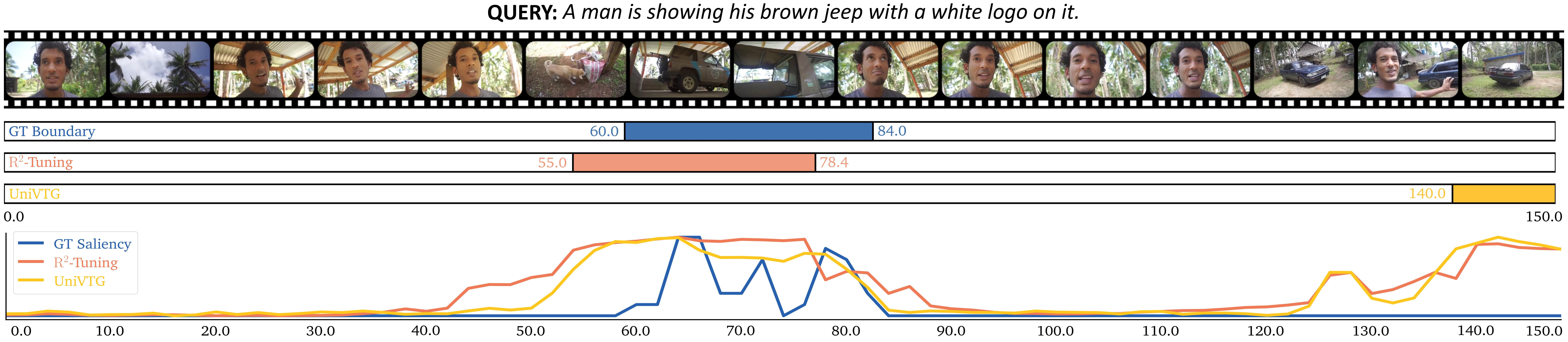}
\vspace{0.1mm}
\end{minipage}
\begin{minipage}{\linewidth}
\includegraphics[width=\textwidth]{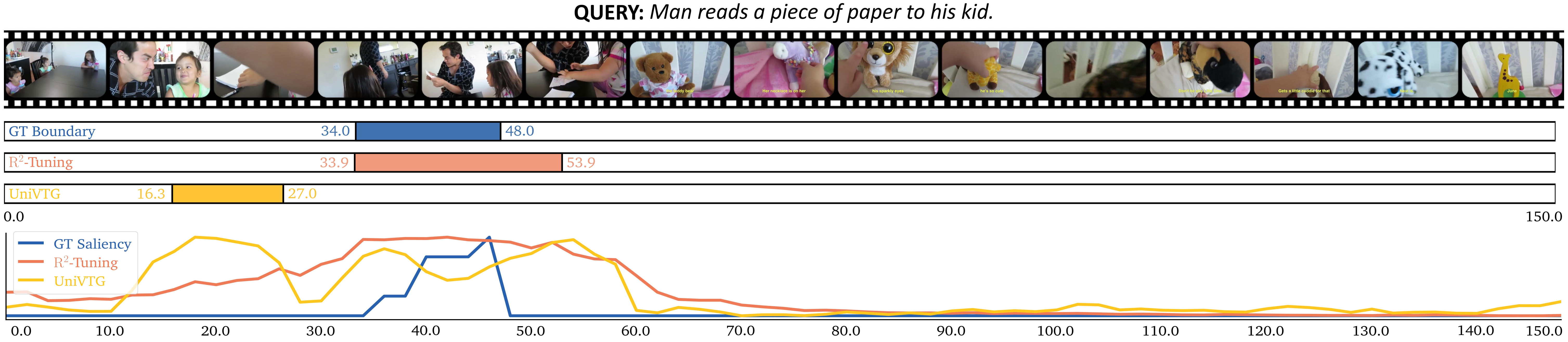}
\vspace{0.1mm}
\end{minipage}
\begin{minipage}{\linewidth}
\includegraphics[width=\textwidth]{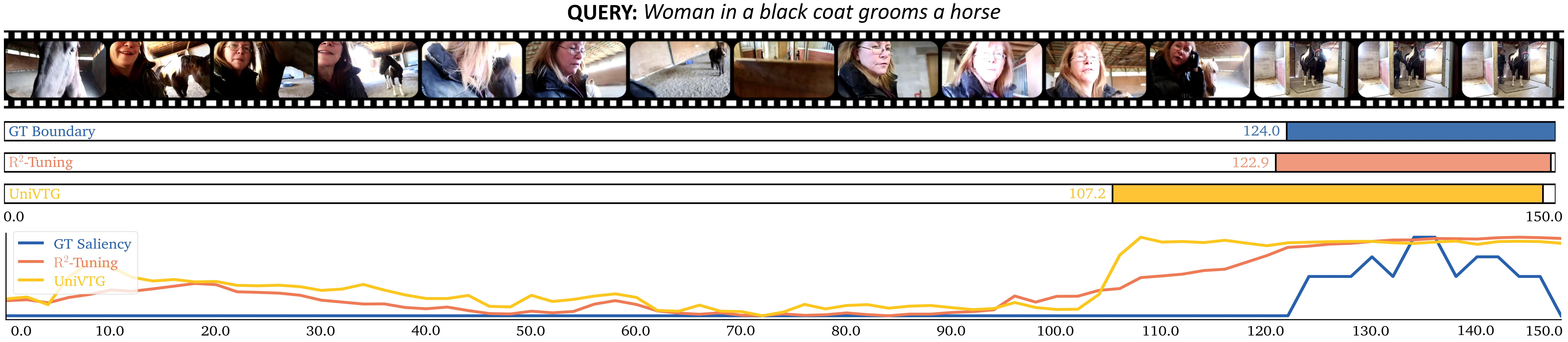}
\vspace{0.1mm}
\end{minipage}
\begin{minipage}{\linewidth}
\includegraphics[width=\textwidth]{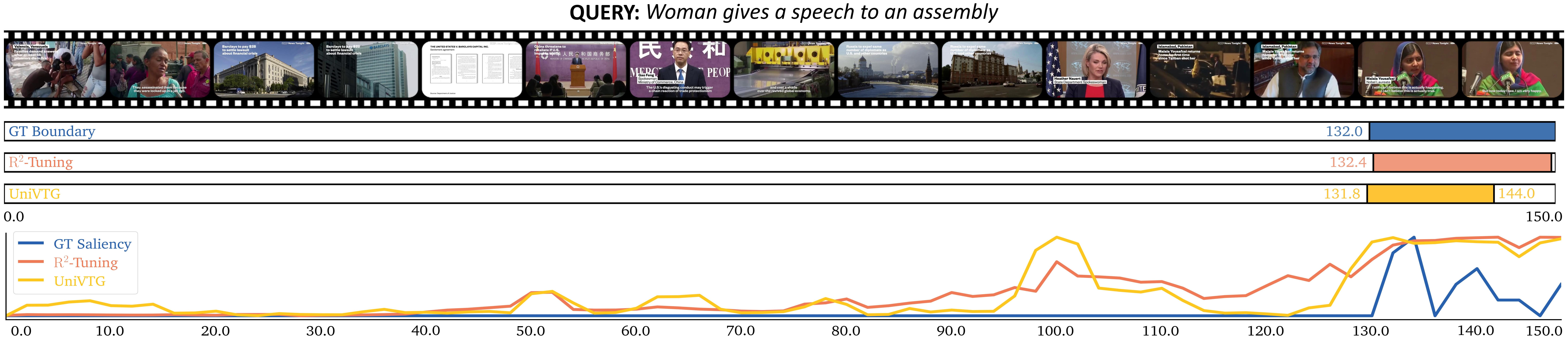}
\vspace{0.1mm}
\end{minipage}
\begin{minipage}{\linewidth}
\includegraphics[width=\textwidth]{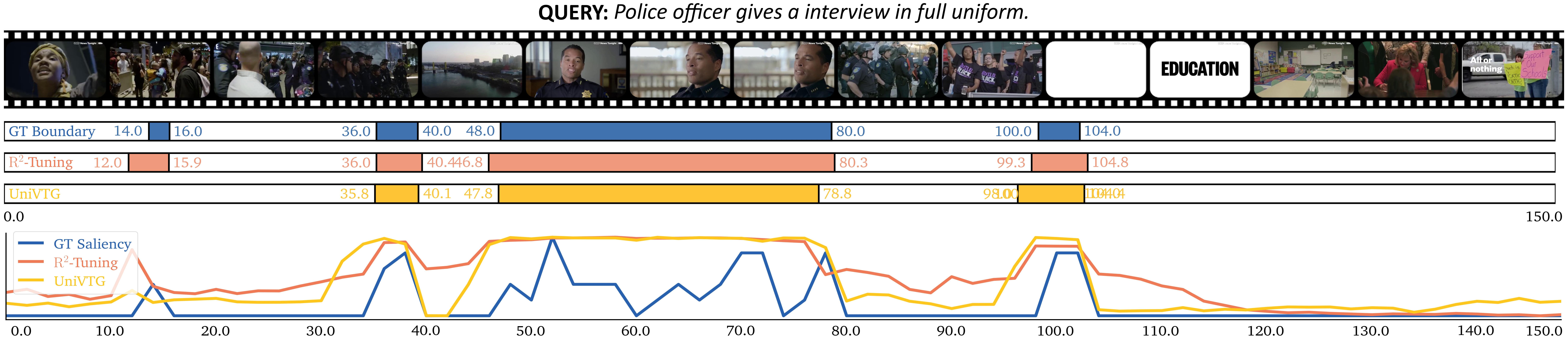}
\end{minipage}
\caption{More visualizations of joint moment retrieval and highlight detection results on QVHighlights \texttt{val} split. \model{} can accurately regress the boundaries of moments and predict highlight saliency scores through its novel feature refinement design.}
\label{fig:more}
\end{figure*}

\begin{figure*}
\centering
\begin{minipage}{\linewidth}
\includegraphics[width=\textwidth]{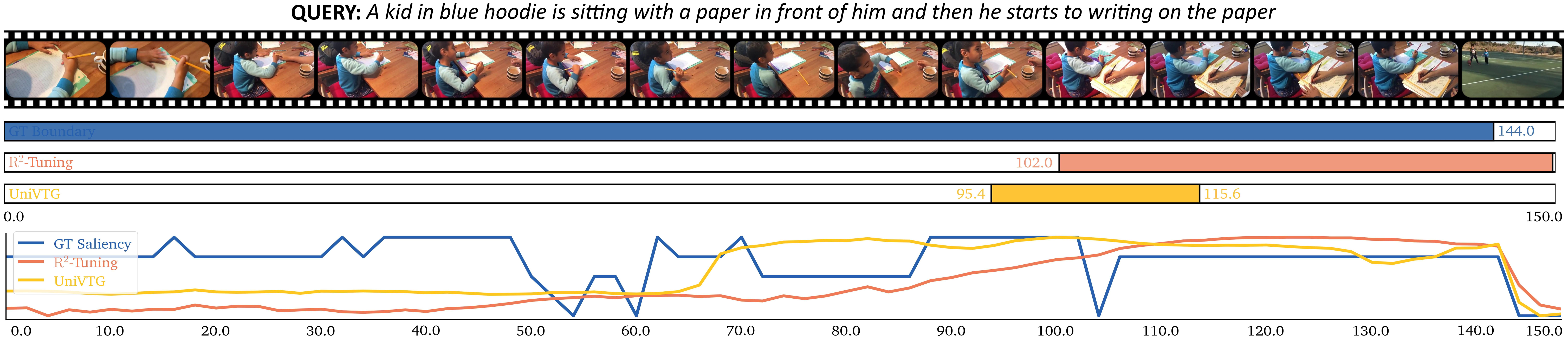}
\vspace{0.1mm}
\end{minipage}
\begin{minipage}{\linewidth}
\includegraphics[width=\textwidth]{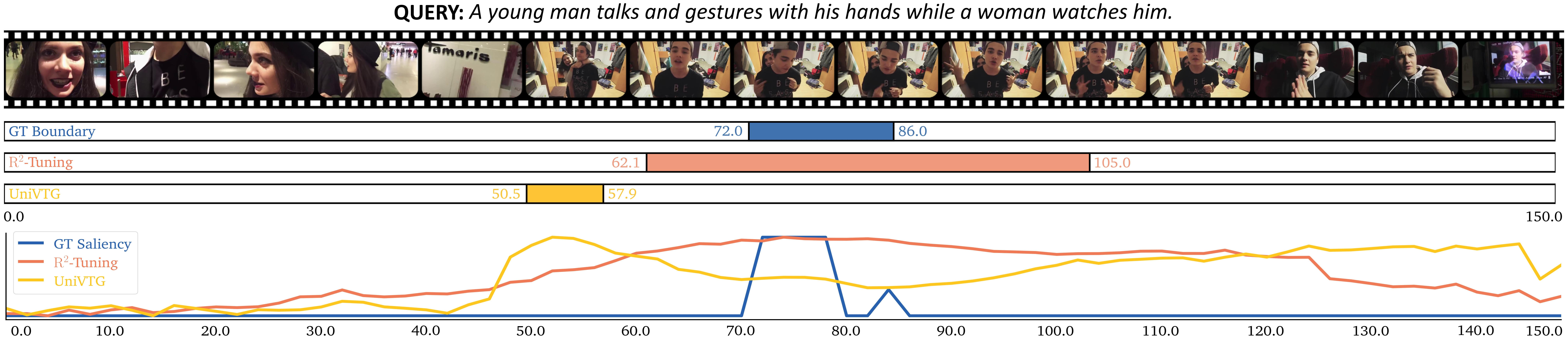}
\end{minipage}
\caption{Failure cases of joint moment retrieval and highlight detection results on QVHighlights \texttt{val} split.}
\label{fig:failure}
\end{figure*}

We show more visualizations of qualitative results in Figure~\ref{fig:more}. Compare with the strong baseline UniVTG \cite{lin2023univtg}. \model{} can regress moment boundaries and detect highlights more accurately due to the novel design. To better study our method, we also show some failure cases in Figure~\ref{fig:failure}. In the first case, when facing complex queries with multiple actions, our method focuses more on a single action (\ie, \textit{writing}). In the second case, we argue that the moment prediction is better than the ground truth.

\section{Limitations \& Future Work}

Our model only considers a single visual modality, while videos naturally contain audio information that can also help detecting moments, highlights, and summaries. Previous works \cite{hong2020mini,badamdorj2021joint,liu2022umt} have demonstrate that audio can largely improve the VTG performances. Thus one of our future works is to incorporate audio into the proposed framework, so that the noise in either modality could be eliminated by the other.

\bibliographystyle{splncs04}
\bibliography{main}

\end{document}